\documentclass[10pt,twocolumn,letterpaper]{article}

\usepackage{cvpr}      
\usepackage{multirow}
\definecolor{cvprblue}{rgb}{0.21,0.49,0.74}
\usepackage[pagebackref,breaklinks,colorlinks,allcolors=cvprblue]{hyperref}

\def\paperID{*****} 
\def\confName{CVPR}
\def\confYear{2026}

\title{\textit{TEA}: Temporal Adaptive Satellite Image Semantic Segmentation}

\author{Juyuan Kang\textsuperscript{1*} \quad Hao Zhu\textsuperscript{1*} \quad Yan Zhu\textsuperscript{1} \quad Wei Zhang\textsuperscript{1}\\
Jianing Chen\textsuperscript{1} \quad Tianxiang Xiao\textsuperscript{1} \quad Yike Ma\textsuperscript{1} \quad Feng Dai\textsuperscript{1†}\\
\textsuperscript{1}Institute of Computing Technology, Chinese Academy of Sciences\\
Beijing, China\\
{\tt\small \{kangjuyuan23s, fdai\}@ict.ac.cn}
}

\begin{document}
\maketitle
\begin{abstract}
\let\thefootnote\relax\footnote{\textsuperscript{*}Equally contributing first authors. 
  \textsuperscript{†}Corresponding author. \\
  Preprint. Under review.}

Crop mapping based on satellite images time-series (SITS) holds substantial economic value in agricultural production settings, in which parcel segmentation is an essential step. 

Existing approaches have achieved notable advancements in SITS segmentation with predetermined sequence lengths. 

However, we found that these approaches overlooked the generalization capability of models across scenarios with varying temporal length, leading to markedly poor segmentation results in such cases.
To address this issue, we propose \textbf{TEA}, a \textbf{TE}mporal \textbf{A}daptive SITS semantic segmentation method to enhance the model's resilience under varying sequence lengths.
We introduce a teacher model that encapsulates the global sequence knowledge to guide a student model with adaptive temporal input lengths. 
Specifically, teacher shapes the student's feature space via intermediate embedding, prototypes and soft label perspectives to realize knowledge transfer, while dynamically aggregating student model to mitigate knowledge forgetting.
Finally, we introduce full-sequence reconstruction as an auxiliary task to further enhance the quality of representations across inputs of varying temporal lengths.

Through extensive experiments, we demonstrate that our method brings remarkable improvements across inputs of different temporal lengths on common benchmarks.

Our code will be publicly available \href{https://github.com/KeplerKang/TEA}{here}.

\end{abstract}    
\section{Introduction}
\label{sec:intro}

\begin{figure}
    \centering
    \includegraphics[width=1\linewidth]{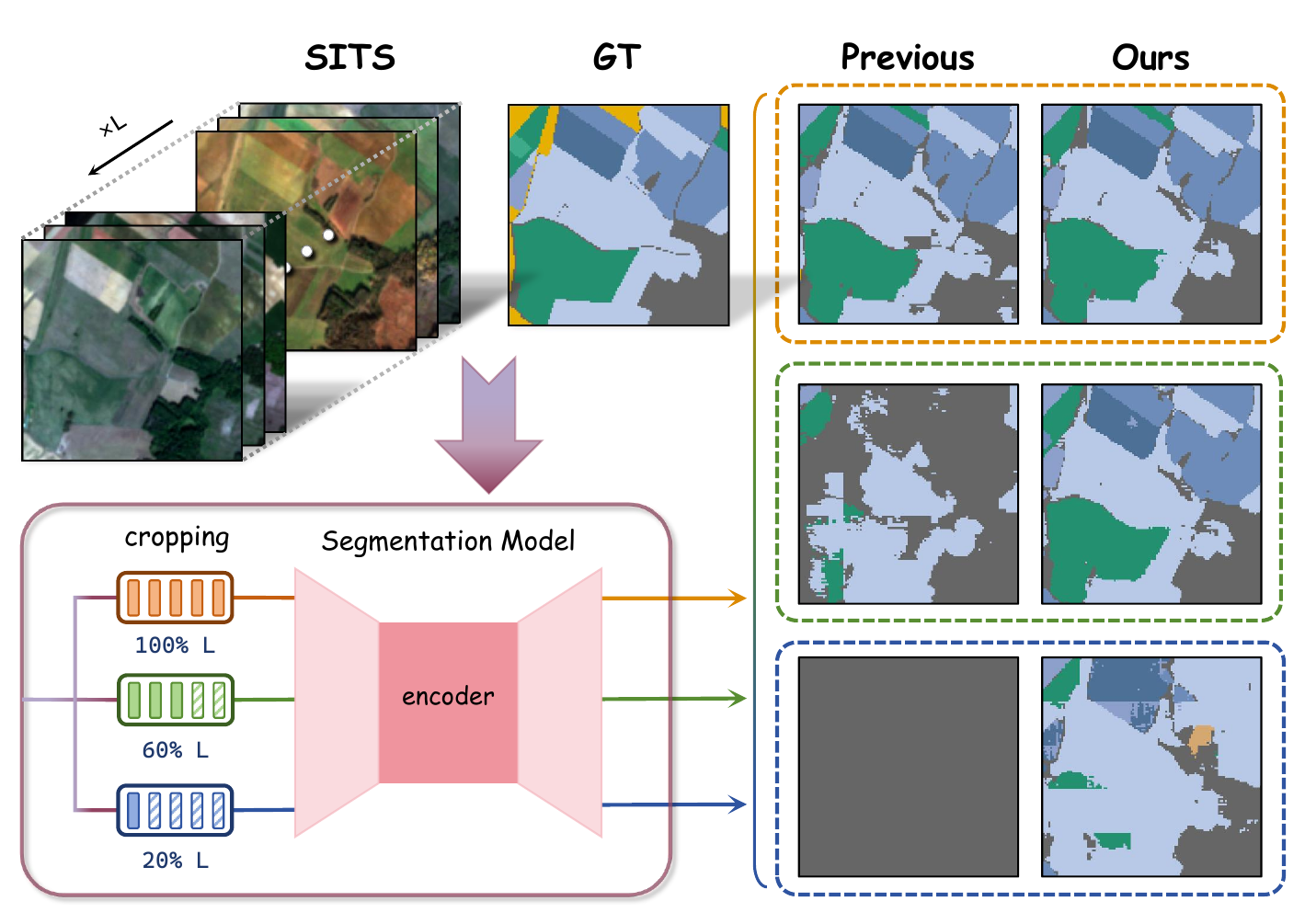}
    \caption{Comparative crop mapping results on \textbf{short sequences SITS}, highlighting prior models in contrast to ours. L denotes the original sequence length. We evaluate segmentation performance on cropped sequences at length ratios of 20\% and 60\% as examples. The results expose shortcomings of prior methods, while showcasing our method’s superiority and robustness across varying temporal lengths.}
    \label{fig:intro}
\end{figure}

Crop mapping \cite{Omnisat, S2, S1, woodcock2008free}facilitated by advanced satellite remote sensing, has substantially advanced agricultural monitoring by enabling large-scale, periodic assessments of crop growth dynamics.

Toward automated crop mapping, a pivotal step is segmenting crop parcels \cite{benson2024multi, Satmae, Skysense, S2mae, improvement, zhong2019deep} from remote sensing imagery, which is essential for downstream analysis.
Satellite Image Time Series (SITS) with multiple spectral channels and high revisit frequency, enable robust characterization of complex temporal variations in crop biophysical properties, making them indispensable for fine-grained crop parcel segmentation. 
In practical agricultural applications, however, predictions often need to be made from a limited number of temporal observations when remote-sensing data over the target area are scarce. 
This situation is common in agricultural remote-sensing monitoring. 
On one side, due to differences in crop growth cycles and crop rotation practices, the same parcel may encompass multiple crop types over a long time span. 
On the other side, timely identification of crop types in the early growth stage provides crucial guidance for downstream decisions and is of great significance for agricultural management, disaster early warning, and precision operations.

At present, extensive work on SITS has driven substantial progress, where numerous models have achieved advanced semantic segmentation performance\cite{FPNConvLSTM, BiConvGRU, BiConvLSTM, ConvGRU, ConvLSTM, TSViT, UNET3D, UNET3Df, UTAE}.
However, most existing models target predetermined lengths SITS, \ie, series spanning more than a calendar year or an entire crop growth cycle, yet they commonly neglect generalization across varying temporal lengths. 
We evaluate the model under varying sequence lengths and notice that, at a temporal length of only 60\% of the origin, segmentation performance degrades markedly; at as short as 20\%, model fails entirely to segment parcels, as shown in Fig.\ref{fig:intro}. 
The results reveal that prior models struggle to accommodate SITS with varying sequence lengths, exhibiting progressively worse performance as the sequences become shorter.
This lack of generalization to temporal length severely limits the applicability of these segmentation models to crop mapping in agricultural settings.

In this paper, we propose a temporal-adaptive SITS semantic segmentation method called \textbf{TEA}, which improves SITS segmentation performance at arbitrary temporal lengths. 
To this end, we first adopt a teacher model that encapsulates global temporal knowledge, guiding the student with adaptive temporal length inputs to learn temporal priors. 
In particular, we leverage the teacher’s spatiotemporal embeddings and soft-label outputs to supervise the student’s feature extraction on short sequences after randomly cropped. 
Futhermore, we design a temporal prototype representation alignment module to enlarge class decision boundaries for short sequences and to enforce consistent representations for the same crop across different growth stages.
Finally, to enhance stability under noisy data, we introduce data reconstruction as an auxiliary task to improve the feature-extraction robustness of the temporal transformer.

To highlight performance under short sequences, we propose the Length-Decayed IoU (LDIoU) metric for a more comprehensive and fair evaluation of our method.

Our main work and contributions are reflected in the following aspects:

\begin{itemize}
    \item We are the first to identify a key limitation of existing SITS segmentation models, \ie, they exhibit strong dependence on temporal length and generalize poorly to short sequences.
\end{itemize}

\begin{itemize}
    \item To handle inputs of varying sequence lengths, we propose TEA, a temporally adaptive segmentation approach that transfers global temporal knowledge and aligns prototype features to counter degraded sensitivity on variable-length sequences, thereby improving generalization and robustness. 
\end{itemize}

\begin{itemize}
    \item Experiments on standard datasets show that our method markedly improves adaptability to varying temporal length SITS. Across multiple ratios of sequence length, it delivers a 19\% absolute gain over the baseline, achieving multi-fold performance improvement. Notably, our approach also surpasses existing models on full-length sequences. Compared with prior methods, it substantially expands the applicability and practical value of temporal remote-sensing segmentation models.
\end{itemize}

\section{Related Work}
\subsection{Semantic Segmentation on SITS.}
Semantic segmentation of agricultural parcels from satellite imagery has attracted widespread attention. The goal is to assign each pixel to the corresponding crop category or background. Traditional remote-sensing mapping methods typically rely on single-date imagery: some approaches perform boundary detection followed by post-processing to obtain segmentation results, while others first use segmentation networks for preprocessing and then compute clustering-based features.

In recent years, parcel segmentation using Satellite Image Time Series (SITS) has gained increasing traction \cite{3DCNN, CNNRNN, TempCNN, GLTAE}. For example, U-ConvLSTM \cite{m2019semantic} encodes spatial features within a U-Net \cite{ronneberger2015u} framework and subsequently encodes temporal features via ConvLSTM \cite{ConvLSTM}; similarly, FPN-ConvLSTM \cite{FPNConvLSTM} replaces U-Net with a Feature Pyramid Network (FPN) \cite{lin2017featurepyramidnetworksobject} as the spatial encoder. U-TAE \cite{UTAE} substitutes the traditional LSTM with a more efficient temporal attention mechanism, enabling better characterization of temporal dynamics. Other methods first extract temporal features and then process the spatial dimension. TSViT \cite{TSViT, zhu2025exact} systematically demonstrates the benefits of prioritizing temporal processing, employing a purely Transformer-based architecture to extract information from SITS and achieving state-of-the-art performance at lower computational cost.

Contrary to most existing methods, our approach places greater emphasis on performance across varying sequence lengths. In agricultural remote-sensing monitoring, due to crop rotation and the importance of early, accurate crop mapping, it is common to require parcel-level semantic segmentation at arbitrary temporal lengths—an aspect often overlooked by existing studies. In this paper, we adopt TSViT’s spatiotemporal encoding as the backbone and propose TEA, a temporally adaptive semantic segmentation model.

\subsection{Short time-series classification.} 
Time-series classification has been researched for long, with several studies targeting early or short-sequence decisions in time-sensitive settings\cite{garnot2020satellite, ghalwash2014utilizing, grabocka2014learning}. 
TEASER\cite{schafer2020teaser} formulates eTSC as a two-tier procedure, enabling earlier decisions without sacrificing accuracy. 
ELECTS\cite{End-to-end}, tailored for crop classification, offers a general end-to-end framework. 
However, these methods focus on early decisions for one single-dimensional time series. 
In contrast, we investigate early decision-making in the context of semantic segmentation on SITS, where data are inherently spatio-temporal.

\begin{figure*}
    \centering
    \includegraphics[width=1.0\linewidth]{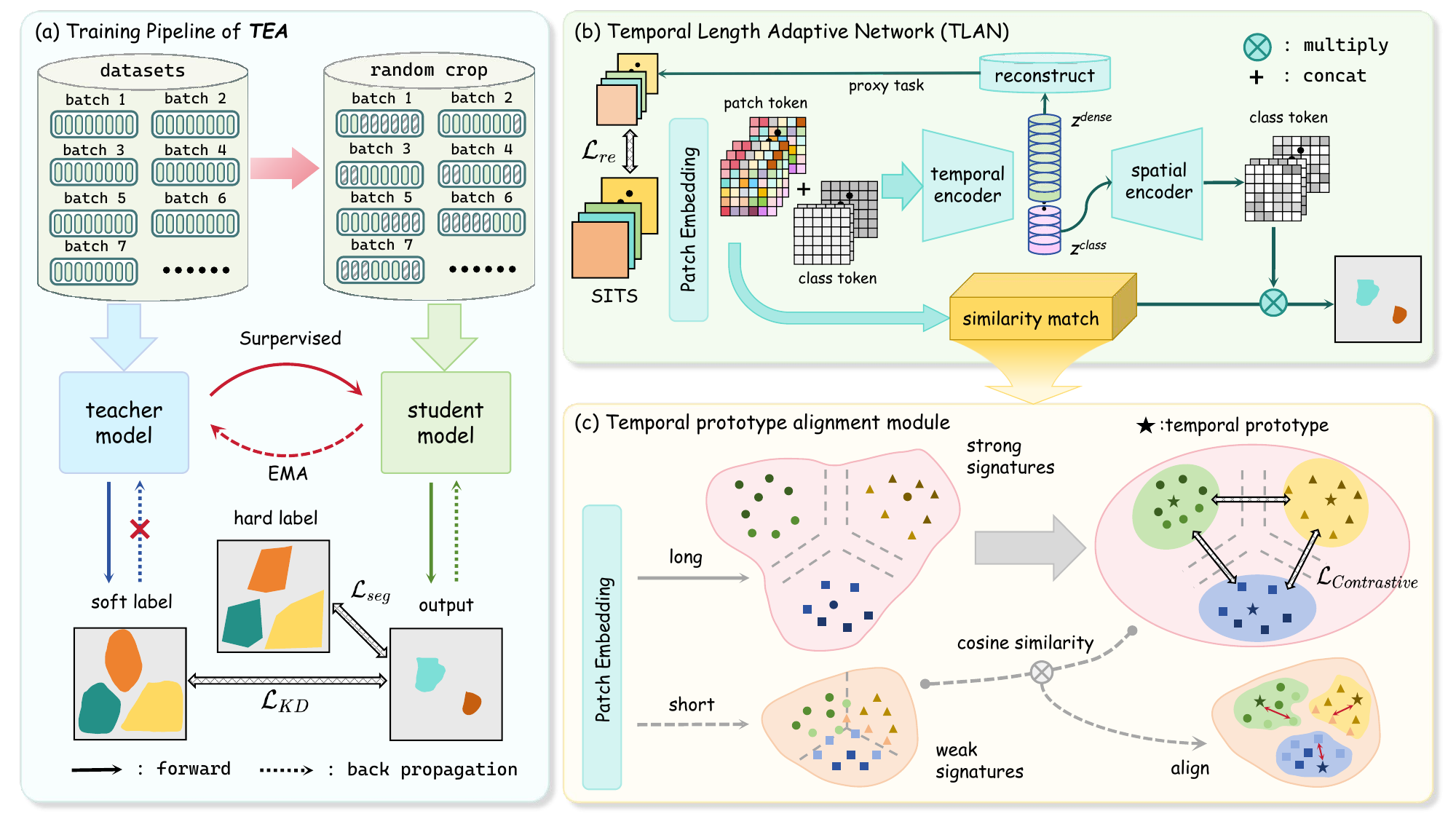}
    \caption{(a)\textbf{The training pipeline of \textit{TEA}.} We adopt a teacher model which captures global temporal representations to supervise a temporal-adaptive student model. (b)\textbf{Our Temporal Length Adaptive Network (TLAN) Insight.} Model leverages the Temporal-Spatio Transformer backbone to process SITS inputs. To improve noise robustness and robustness across varying sequence lengths, we introduce an auxiliary task, data reconstruction, alongside a temporal prototype alignment mechanism. (c)\textbf{Temporal prototype alignment module} expand decision boundaries via clustering to enhance class discriminability for categories with weak temporal signatures. }
    \label{fig:method}
\end{figure*}

\subsection{Knowledge Distillation.} 
Knowledge distillation has been extensively studied for a long time \cite{ahn2019variational, aguilar2020knowledge, bergmann2020uninformed, allen2019learning}, tracing back to early research in model compression \cite{buciluǎ2006model}.
In 2015, Hinton et al. \cite{hinton2015distilling} systematized this idea as “knowledge distillation,” introducing temperature scaling and a joint loss, and generalizing it into a universal teacher–student learning paradigm.
Tarvainen et al. \cite{tarvainen2017mean} proposed the Mean Teacher method in 2017, which averages model weights via Exponential Moving Average (EMA) instead of label predictions.
Wang et al. \cite{wang2018dataset} proposed dataset distillation in 2018, aiming to replace the original large-scale dataset with a very small set of “synthetic representative data”.
Rather than pursuing model lightening or data reduction, we use a teacher–student framework to distill global temporal knowledge into a student that learns robustly across variable sequence lengths.

\section{Method}
In this section, we first introduce our task definition and the feature-extraction backbone used (\cref{preliminaries}). 
We then describe the training pipeline (\cref{subsec:pipeline}) for a sequence-length-adaptive segmentation model under a teacher–student knowledge distillation framework. 
Next, we detail the internal design of the model: (1) mining class-aware cues from short sequences and injecting them into the model’s feature-layer outputs via confidence accumulation (\cref{subsec:temporal_prototype}), and (2) introducing an auxiliary task of data reconstruction (\cref{subsec:reconstruct}) to enhance the model’s feature extraction capability. Finally, we propose an evaluation metric LDIoU (\cref{ldiou}) tailored for short temporal sequences, facilitating quantitative comparisons in future related studies.

\subsection{Preliminaries}
\label{preliminaries}
The semantic segmentation of remote sensing image series has developed a well-established technical paradigm.
They model the Satellite Image Time Series (SITS) input as a tensor $\mathbf{X} \in \mathbb{R}^{T \times C \times H \times W}$, where $T$ denotes the sequence length of the SITS data, $C$ represents the number of spectral channels, and $H \times W$ corresponds to the spatial resolution of the input images. 
However, prior works have focused solely on segmentation performance at the full sequence length T, leading to our task: adaptive-length temporal image segmentation..  

Among numerous existing models for temporal image segmentation, modeling temporal dynamics upfront and then extracting spatial features yields strong performance on SITS. 
Following the design of TSViT, we use a Vision Transformer variant as the backbone for both the temporal and spatial encoders. 

First, we partition $\mathbf{X}$ into non-overlapping spatial patches and convert these patches into a sequence of patch tokens. This tokenization process reshapes the input into $\mathbf{Z} \in \mathbb{R}^{N_h \cdot N_w \times T \times d}$, where $N_h = \lfloor \frac{H}{h} \rfloor$ and $N_w = \lfloor \frac{W}{w} \rfloor$ (here, $h \times w$ denotes the spatial size of each individual patch).

Subsequently, temporal positional embeddings $\mathbf{P}_T \in \mathbb{R}^{T \times d}$ are incorporated into the temporal dimension of $\mathbf{Z}$, and the augmented tensor is concatenated with temporal multi-class tokens $\mathbf{Z}_T^{\text{cls}} \in \mathbb{R}^{K \times d}$ to form the input for the temporal encoder. The resulting input tensor is defined as:
\begin{equation}
    \mathbf{Z}_T^{\text{in}} = \textrm{concat}( \mathbf{Z}_T^{\text{cls}}, \mathbf{Z} + \mathbf{P}_T ), \mathbf{Z}_T^{\text{in}} \in \mathbb{R}^{N_h \cdot N_w \times (K + T) \times d}
\end{equation}

In the above equation, $K$ stands for the number of target categories. To align with the spatial grid structure of $\mathbf{Z}$, both $\mathbf{Z}_T^{\text{cls}}$ and $\mathbf{P}_T$ are replicated $N_h \cdot N_w$ times.

From the output of the temporal encoder $\mathbf{Z}_T^{\text{out}}$, we extract the first $K$ tokens to obtain dense temporal tokens $\mathbf{Z}_T^{\text{dense}} \in \mathbb{R}^{N_h \cdot N_w \times K \times d}$. We then swap the first two dimensions of $\mathbf{Z}_T^{\text{dense}}$ to construct spatial input tokens $\mathbf{Z}_S \in \mathbb{R}^{K \times N_h \cdot N_w \times d}$, which serve as the input for the spatial encoder. For each of the $K$ spatial streams, these tokens are further enhanced with spatial multi-class tokens $\mathbf{Z}_S^{\text{cls}} \in \mathbb{R}^{K \times 1 \times d}$ and spatial positional embeddings $\mathbf{P}_S \in \mathbb{R}^{N_h \cdot N_w \times d}$. The input to the spatial encoder is thus formulated as:
\begin{equation}
    \mathbf{Z}_S^{\text{in}} = \textrm{concat}( \mathbf{Z}_S^{\text{cls}}, \mathbf{Z}_S + \mathbf{P}_S ), \mathbf{Z}_S^{\text{in}} \in \mathbb{R}^{K \times N_h \cdot N_w \times d}
\end{equation}

After the spatial encoding stage, the output features $\mathbf{Z}_S^{\text{out}}$ are split into two components: $\mathbf{Z}_S^{\text{out}} = \left[ \mathbf{Z}_S^{\text{global}}, \mathbf{Z}_S^{\text{dense}} \right]$, which are tailored for different downstream tasks. For the classification task, the global tokens $\mathbf{Z}_S^{\text{global}} \in \mathbb{R}^{K \times 1 \times d}$ are fed into a classifier to generate category logits. For the segmentation task, the dense tokens $\mathbf{Z}_S^{\text{dense}} \in \mathbb{R}^{K \times N_h \cdot N_w \times d}$ are passed to a segmentation decoder, which produces dense prediction masks corresponding to the input spatial resolution.

\subsection{Training pipeline}
\label{subsec:pipeline}
\subsubsection{Random Crop}
 In previous parcel segmentation tasks based on temporal remote sensing images, a complete set of Satellite Image Time Series (SITS) is required as input, denoted as \( X \in \mathbb{R}^{T \times C \times H \times W} \). Here, \( T \) represents the length of the time series—specifically, "complete" in this context means \( T \) typically corresponds to a time span of more than one year; \( C \) denotes the number of channels of the remote sensing images; and \( H \times W \) refers to the spatial dimensions (i.e., height \( H \) and width \( W \)) of the images. However, our task aims to achieve parcel segmentation using time series of shorter length, which can be adaptive to various sequence lengths. 
 
 To this end, we perform masking along the temporal dimension, crop the original complete SITS samples into subsequences, denoted as \( X' \in \mathbb{R}^{T \times C \times H \times W} \). with the shorter length. 
 For the training set, we use data with random lengths masked at random positions as input to train the model, enabling the model to fully learn features from all parts of the sequence. 
 For the validation and test sets, to stably obtain subsequences at the same position, we perform cropping starting from the first frame, while adopting ten graduated ratio levels (starting at 10\% and ending at 100\%), denoted as $\text{ratio} = \{10\%, \dots, 100\%\}$. We crop data from the start of the sequence according to these ratios to obtain ten types of early-stage sequences with different lengths as model inputs—this design is motivated by the practical requirement of parcel segmentation based on early-stage images of crop growth.

\subsubsection{Knowledge Distillation}
As a classic paradigm for knowledge distillation, the teacher-student framework is commonly employed for knowledge transfer and model lightweighting. For the early-stage sequence segmentation task, we utilize the complete temporal data features to serve as the teacher (model), which is used to optimize the features extracted by the student model from short temporal data. This design aims to enhance the student model’s robustness to temporal data of different lengths and its ability to perceive global parcels.

We conduct knowledge transfer on three distinct components: the embeddings generated by the temporal and spatial encoders, the temporal prototype representations (elaborated in \cref{subsec:temporal_prototype}), and the soft labels. Specifically, cross-entropy loss is utilized to supervise the student model in aligning its global temporal features with those of the teacher model within the fine-grained feature space.

Specifically, We define the temporal feature loss \( \mathcal{L}_t \) (average MSE loss) to measure the discrepancy between the temporal features of the student and teacher models. Let \( F_t^T \in \mathbb{R}^{K \times d} \) and \( F_s^T \in \mathbb{R}^{K \times d} \) denote the temporal features extracted from the teacher and student networks, respectively. Then:
\begin{equation}
    \mathcal{L}_t = \frac{\sum\limits_{n=1}^N \sum\limits_{k=1}^K \sum\limits_{d=1}^D \left( F_t^T[n][k][d] - F_s^T[n][k][d] \right)^2}{N \cdot K \cdot D} 
\end{equation}

Similarly, we use the MSE loss to calculate the discrepancy between the spatial features of the student and teacher models, defined as \( \mathcal{L}_s \). Let \( F_t^S \in \mathbb{R}^{K \times (N_h \cdot N_w + 1) \times d} \) and \( F_s^S \in \mathbb{R}^{K \times (N_h \cdot N_w + 1) \times d} \) represent the spatial features of the teacher and student networks, respectively. Then:

\begin{equation}
    \mathcal{L}_s = \frac{ \sum\limits_{n=1}^{N_h \cdot N_w } \sum\limits_{k=1}^{K} \sum\limits_{d=1}^D \left( F_t^S[n][k][d] - F_s^S[n][k][d] \right)^2}{N_h \cdot N_w \cdot K \cdot D}
\end{equation}

\subsubsection{Exponential Weighted Moving Average}
Teacher model updates its parameters incrementally using the Exponential Moving Average (EMA) method, instead of backpropagation. The parameter update strategy is formulated as:
\begin{equation}
    \text{Teacher} = \text{teacher} \cdot \text{decay} + (1 - \text{decay}) \cdot \text{student}
\end{equation}
where \( \text{Tea} \) denotes the parameters of teacher, \( \text{tea} \) represents the parameters of the teacher model before update, \( \text{stu} \) refers to the parameters of the student model, and \( \text{decay} \in \{\text{linear}, \text{exponential}\} \) , i.e., the decay rate adopt either a linear schedule to warmup before a exponential schedule.

\subsection{Data reconstruction}
\label{subsec:reconstruct}
Within the aforementioned long-short temporal teacher-student distillation framework, to enhance the robustness of the temporal encoder in fitting features of temporal data under window masking at different positions, we introduce an auxiliary task: reconstructing the original data from the output of the temporal encoder.

For an input data sample \( z_{\text{in}} = x_{T_s,r} \) (where \( T_s \) denotes the length of the short temporal sequence, and \( r \) is the cropping ratio), consider the output of the temporal encoder \( Z_{\text{out}}^T = [Z_{\text{dense}}^T \mid Z_{\text{seq}}^T] \) (where \( \mid \) denotes concatenation along the feature dimension). We extract its feature component \( z_{\text{dense}} \), specifically:
\begin{equation}
    Z_t^{\text{seq}} \in \mathbb{R}^{N \times N \times T \times d}
\end{equation}
where \( N = N_h = N_w \) (assuming square patches for simplicity, consistent with \( N_h = \lfloor \frac{H}{h} \rfloor \) and \( N_w = \lfloor \frac{W}{w} \rfloor \) defined earlier), \( T \) is the temporal length of the encoded features, and \( d \) is the feature dimension.

Next, we design a temporal feature decoder \( D \) aimed at deconstructing \( z_t \) into a reconstruction output with dimensions equivalent to the original input data, denoted as \( z_{\text{out}}^{\text{seq}} = x_{T_s}' \in \mathbb{R}^{H \times W \times T_s \times C} \). Specifically, we adopt a structure corresponding to the encoder: transposed convolution operations are applied along the feature dimension to obtain data with dimensions \( \text{patchsize}^{2} \times \text{channels} \), followed by dimension rearrangement to adjust to the same dimension as the input data. 

We then use the MSE loss to measure the difference between the original data and the reconstructed data, defined as:
\begin{equation}
    \mathcal{L}_{\text{reconstruct}} = \frac{\sum\limits_{n=1}^{H \times W} \sum\limits_{t=1}^{T_s} \sum\limits_{c=1}^C \left( x_{T_s,r}[n][t][c] - z_{\text{out}}^{\text{seq}}[n][t][c] \right)^2}{H \cdot W \cdot T_s \cdot C} 
    \label{eq:important}
\end{equation}

where \( x_{T_s,r} \) represents the original input short temporal sequence, and \( z_{\text{out}}^{\text{seq}} \) denotes the reconstructed output.

\begin{table*}
    \caption{Various temporal length performance of different models on PASTIS \cite{UTAE} and Germany \cite{germany}. Baseline refers to TSViT\cite{TSViT}, which is the backbone of ours method. The metric adopts mIoU as the evaluation criterion. We both report the standard mean mIoU (mIoU) and our proposed Length-Decayed mIoU (LDIoU). The best results are highlighted in \textbf{bold}, while the second-best results are marked in \underline{underline}. (* denotes the introduction of an FPN \cite{lin2017featurepyramidnetworksobject} head.)}
    \centering
    \resizebox{\textwidth}{!}{%
    \begin{tabular}{*{1}{c}|*{1}{c}|*{10}{c}|*{1}{c}|*{1}{c}}
        \toprule[1pt] 
        \multirow{2}{*}{ }  & \multirow{2}{*}{Method}   & \multicolumn{10}{c|}{ratios}  &mmIoU  & LDIoU   \\
                    & &10\% &20\% &30\% &40\% &50\% &60\% &70\% &80\% &90\% &100\% &(\%) &(\%) \\
        
        \midrule 
        \multirow{10}{*}{\rotatebox{90}{\textit{\textbf{PASTIS }}\cite{UTAE}}}  
        &UNet3D \cite{UNET3D} & 3.23 & 4.17 & 4.77 & 5.86 & 6.34 & 11.18 & 19.83 & 39.14 & 47.32 & 50.84 & 19.27 & 10.10 \\
        &UNet3Df \cite{UNET3Df} & 3.63 & 3.72 & 3.74 & 4.95 & 6.91 & 9.66 & 20.58 & 33.50 & 44.59 & 49.58 & 18.09 & 9.56 \\
        &ConvGRU \cite{ConvGRU}& 2.40 & 3.94 & 4.28 & 4.82 & 5.36 & 7.26 & 7.24 & 15.59 & 28.74 & 50.57 & 13.02 & 7.00 \\
        &BiConvGRU  \cite{BiConvGRU}&2.29 &3.83 &4.13 &5.38 &7.45 &9.42 &15.59 &25.15 &24.39 &46.42 &14.40 &7.75  \\
        &ConvLSTM \cite{ConvLSTM}& 3.28 & 3.34 & 3.47 & 3.58 & 3.89 & 4.23 & 5.17 & 7.85 & 22.11 & 51.67 & 10.86 & 6.09 \\
        &ConvLSTM* \cite{FPNConvLSTM}& 0.36 & 0.38 & 0.53 & 1.06 & 1.45 & 0.74 & 0.55 & 2.00 & 29.00 & 55.90 & 9.20 & 3.60 \\
        &BiConvLSTM \cite{BiConvLSTM}  &\underline{5.51} &\underline{6.19} &\underline{7.53} &\underline{9.07} &9.31 &9.85 &13.78 &21.84 &29.72 &46.50 &15.93 &10.08  \\
        &U-TAE \cite{UTAE}& 3.15 & 5.93 & 6.34 & 8.67 & 11.19 & \underline{21.52} & 33.17 & 52.72 & 60.55 & 61.40 & 26.46 & 13.80 \\
        &baseline & 3.81 & 5.60 & 6.10 & 6.48 &\underline{11.20} &20.34 &\underline{34.42} &\underline{56.46} &\underline{62.65} &\underline{64.08} &\underline{27.11} &\underline{14.08} \\
        &TEA(ours) &\textbf{21.5} &\textbf{26.22} &\textbf{28.43} &\textbf{32.70} &\textbf{37.57} &\textbf{45.82} &\textbf{56.45} &\textbf{65.36} &\textbf{66.37} &\textbf{66.77} &\textbf{44.72} &\textbf{33.36} \\
        \midrule 
        \multirow{10}{*}{\rotatebox{90}{\textit{\textbf{Germany }}\cite{germany}}} 
        &UNet3D & 2.44 & 2.48 & 2.43 & 3.01 & 11.94 & 16.93 & 27.05 & 32.52 & 65.79 & 73.66 & 23.83 & 11.29 \\
        &UNet3Df & \textbf{2.65} & 3.12 & 3.57 & 3.42 & 15.19 & 15.56 & 30.04 & 39.73 & 65.44 & 74.32 & 25.30 & 12.24 \\
        &ConvGRU & 1.32 & 2.43 & 2.32 & 2.29 & 3.04 & 3.09 & 4.80 & 5.03 & 11.20 & 59.00 & 9.45 & 4.60 \\
        &BiConvGRU   &2.22 &3.03 &3.01 &3.09 &4.03 &3.61 &3.15 &4.67 &6.7 &54.99 &8.85 &4.85  \\
        &ConvLSTM & 0.62 & 1.26 & 1.45 & 1.34 & 3.08 & 2.16 & 6.39 & 10.57 & 26.68 & 63.15 & 11.67 & 4.97 \\
        &ConvLSTM* & 0.58 & 0.53 & 0.58 & 1.93 & 14.64 & 17.41 & 27.65 & 41.85 & 64.60 & 73.19 & 24.30 & 10.59 \\
        &BiConvLSTM  &2.03 &2.16 &1.83 &1.14 &1.49 &1.34 &1.28 &2.49 &6.89 &58.1 &7.88 &3.96  \\
        &U-TAE & 2.44 & 3.37 & 3.26 & \underline{7.69} & 20.80 & 26.30 & 56.55 & 75.87 & 79.39 & 82.00 & 35.77 & 17.16 \\
        &baseline &2.42 &\underline{4.62} &\underline{4.90} &5.55 &\underline{24.74} &\underline{30.58} &\underline{67.57} &\underline{79.22} &\underline{84.30} &\underline{86.28} &\underline{39.02} & \underline{18.90} \\
        &TEA(ours) &\underline{2.49} &\textbf{30.25} &\textbf{34.64} &\textbf{46.92} &\textbf{66.87} &\textbf{72.20} &\textbf{84.18} &\textbf{85.69} &\textbf{86.24} &\textbf{86.36} &\textbf{59.58} &\textbf{36.62} \\
        \bottomrule[1pt] 
    \end{tabular}
    }
    \label{tab:main}
\end{table*}

\subsection{Temporal prototype alignment module}
\label{subsec:temporal_prototype}
We find that the model still lacks the capability to derive segmentation masks from input data when dealing with short window sequences. This is primarily because the temporal encoder fails to acquire sufficient category-specific feature information from short valid temporal windows, which in turn limits the feature extraction capabilities of the spatial encoder and segmentation decoder. Therefore, building upon the original temporal-spatial sequential encoder-decoder structure, we introduce a matrix representing the temporal feature prototypes. By computing its similarity with the input data as the confidence for short temporal sequence classification, we superimpose this confidence onto the output of the spatial encoder. This idea is analogous to prototype learning \cite{yarats2021reinforcement, li2021adaptive, sun2024learning, yang2018robust} ; however, what we need to compare is feature category matching under the scenario of misaligned feature dimensions.

Specifically, we maintain a set of learnable prototype vectors \( P \in \mathbb{R}^{K \times T_p \times D} \), where \( K \) denotes the number of categories, \( D \) is the same dimension as the feature embedding (consistent with the \( d \) defined in Section 3.3), and \( T_p \) represents the temporal dimension of the prototype vectors (set to match the maximum possible length of short temporal sequences in practice). 

First, we expand \( P \) to a dimension equivalent to the number of patches (i.e., \( N_h \cdot N_w \), where \( N_h = \lfloor \frac{H}{h} \rfloor \) and \( N_w = \lfloor \frac{W}{w} \rfloor \) denote the number of patches along the height and width directions, respectively) to facilitate similarity calculation with the encoded features. The expanded prototype vectors are denoted as \( \tilde{P} \in \mathbb{R}^{K \times T_p \times D \times N_h \cdot N_w} \).

We use cosine similarity to compute the similarity between the encoded features and the prototype vectors. Let the encoded short temporal features be \( Z_{\text{enc}} \in \mathbb{R}^{N_h \cdot N_w \times T_s \times D} \) (where \( T_s \) is the length of the input short temporal sequence). The cosine similarity between \( Z_{\text{enc}} \) and \( \tilde{P} \) is calculated as follows:
\begin{equation}
    \text{Cos}(Z_{\text{enc}}, \tilde{P}) = \frac{Z_{\text{enc}} \cdot \tilde{P}}{\| Z_{\text{enc}} \|_2 \cdot \| \tilde{P} \|_2 + \epsilon}
\end{equation}
where \( \cdot \) denotes the dot product operation, \( \| \cdot \|_2 \) represents the L2 norm, and \( \epsilon \) is a small constant (e.g., \( 10^{-6} \)) added to avoid division by zero.

For the resulting cosine similarity matrix, we first take the average over the feature dimension (i.e., the \( D \)-dimension) and then take the average over the temporal dimension. Finally, we obtain a similarity matrix of size \( N_h \cdot N_w \times K \) , where \( N_h \cdot N_w \) is the total number of patches. We directly apply this matrix to the final output of the spatial encoder, i.e., element-wise multiplication or addition.

\subsection{Length-Decayed Weighted mIoU}
\label{ldiou}
Targeting the specific requirements of the varing-length image sequences segmentation task, we propose an evaluation metric tailored to assess the model’s segmentation performance on temporal remote sensing images up to different lengths. 
Since our objective is to achieve accurate segmentation using sequences that are as short as possible, yet models typically degrade in performance as sequence length decreases, the conventional mMIoU metric is ill-suited to emphasizing gains on short temporal sequences.

Based on this insight, we propose using the mean Intersection over Union (MIoU) at each ratio as the base, and normalizing the reciprocals of sequence lengths to obtain a weight sequence that decays with increasing temporal length. 
We then compute the weighted average of MIoU metrics across all ratios using this decaying weight sequence, resulting in a task-specific metric, denoted as LDIoU (Length-Decayed mIoU):
\begin{equation}
\text{LDIoU} = \sum_{\tau_j} \omega_{\tau_j} \cdot \text{MIoU}_{\tau_j}
\label{eq:tdiou}
\end{equation}
where $\text{MIoU}_{\tau_j}$ represents the MIoU at the SITS with sequence length $\tau_j$ ( $\tau_j$ denotes the sequence length), and $\omega_{\tau_j}$ is the weight for length $\tau_j$, calculated as:
\begin{equation}
\omega_{\tau_j} = \frac{1/\tau_j}{\sum_{\tau_j} 1/\tau_j}
\label{eq:tdiou_weight}
\end{equation}
Here, $\tau_j$ iterates over all sequence lengths of the evaluated stages, and the normalization function ensures the weights sum to 1.

\section{Experiments}

\begin{table*}
    \caption{We conduct ablation studies on PASTIS by incrementally adding modules. \textbf{RC} denotes random cropping; \textbf{KD} denotes the teacher–student knowledge distillation framework; \textbf{Re} denotes the data reconstruction task; \textbf{Pro} denotes the temporal prototype alignment module. \textbf{All the proposed training methods and modules are effective.}}
    \centering
    \resizebox{\textwidth}{!}{%
    \begin{tabular}{*{1}{c}|*{4}{c}|*{10}{c}|*{1}{c}|*{1}{c}}
        \toprule[1pt] 
        \multirow{2}{*}{Method} &\multirow{2}{*}{RC} &\multirow{2}{*}{KD} &\multirow{2}{*}{Re} &\multirow{2}{*}{Pro}  &\multicolumn{10}{c|}{ratios}  &mmIoU  & LDIoU \\
         & & & & &10\% &20\% &30\% &40\% &50\% &60\% &70\% &80\% &90\% &100\% &(\%) &(\%) \\
         \midrule
         baseline & & & &   &3.81 &5.60 &6.10 & 6.48 &11.20 &20.34 &34.42 &56.46 &62.65 &64.08 &27.11 &14.08  \\
            &\checkmark & & &   &15.21 &21.30 &24.87 &29.32 &35.30 &43.07 &52.91 &62.30 &64.53 &65.08 &41.39 &28.93 \\
            &\checkmark &\checkmark & &   &21.16 &25.99 &28.61 &32.83 &37.92 &44.88 &55.07 &63.88 &64.96 &65.33 &44.06 &32.97 \\
            &\checkmark & &\checkmark &\checkmark    &21.61 &25.81 &27.54 &32.40 &37.53 &45.45 &55.02 &64.17 &65.19 &65.55 &44.03 &32.97 \\
        ours-TEA &\checkmark &\checkmark &\checkmark &\checkmark  &21.50 &26.22 &28.43 &32.70 &37.57 &45.82 &56.45 &65.36 &66.37 &66.77 &44.72 &33.36 \\
            &\checkmark &\checkmark &\checkmark &    &21.25 &25.22 &27.18 &31.19 &37.07 &44.76 &54.64 &64.09 &65.51 &65.67 &43.66 &32.52 \\
            &\checkmark &\checkmark  & &\checkmark    &20.87 &25.79 &28.64 &32.83 &38.46 &46.00 &55.34 &63.90 &65.78 &65.97 &44.36 &33.01 \\
        \bottomrule[1pt] 
    \end{tabular}
    }
    \label{tab:ablation}
\end{table*}

\subsection{Experimental Setup}
\textbf{Datasets \& evaluation metric.} 
We conduct comprehensive experiments on two widely used Satellite Image Time Series (SITS) datasets to quantitatively verify the effectiveness of our method. The PASTIS \cite{UTAE} dataset consists of 2433 multispectral time series images acquired by Sentinel-2 satellites. Each sequence contains 33 to 61 observations and 10 spectral bands, and it covers 18 crop types plus one background class. Following the setup in \cite{TSViT}, we use fold-1 among the five predefined folds provided in PASTIS for our experiments. The German \cite{germany} dataset comprises 137k time series images with 13 spectral bands. Each sequence in this dataset includes 15 to 45 observations and is labeled into 17 crop types. Each dataset is split into a training set, a validation set, and a test set at a ratio of 3:1:1.
For evaluation metrics, we use Mean Intersection over Union ($\text{mIoU}$) to evaluate model performance at each individual test-length-ratio, and then we adopt mean mIoU and our proposed Length-Decayed mIoU to comprehensively evaluate model performance.
We use these two datasets for experimental validation, with the above metrics serving as the core evaluation criteria.

\textbf{Implementation details.}
For all experiments, our segmentation model was trained on 8 NVIDIA A100 GPUs (40 GB each). Training was conducted for 100 epochs with a batch size of 128, and a global step count was set to record the number of trained batches. For the student model, the Adam optimizer \cite{loshchilov2017decoupled} was used: a linearly increasing learning rate (from 1e-8 to 1e-3) was adopted during the first 10 epochs of the warm-up period, followed by cosine weight decay \cite{loshchilov2016sgdr}. For the teacher model, parameters were updated via the EMA (Exponential Moving Average) method based on steps. The alpha (weight for teacher parameter update) included a warm-up period: it linearly increased from 0.1 to 0.9 within the first 15\% of steps to enable rapid feature learning, and then slowly increased exponentially to 0.999 to acquire more stable knowledge. 
The parameter settings for our baseline model were consistent with those used in \cite{TSViT}. 
Validation was performed every 500 steps, and the model with optimal performance was saved. 
During test, we truncate each sequence starting from the first frame at ten predefined ratios, and compute the IoU for each. Then we comprehensively presented its segmentation performance (including mmIoU , and the proposed LDIoU) on sequences of different lengths, as well as ablation studies for each module. Overall, our settings achieved optimality for the current methods.

\begin{table}
    \caption{\textbf{Impact of different cropping strategies on temporal robustness.} \textbf{R-Ratio} stands for intercepting at a random ratio (from the start), while \textbf{R-Start} stands for intercepting from an arbitrary position. \textbf{Both} significantly enhances the model’s adaptability to varying sequence lengths.}
    \centering
    \resizebox{0.48\textwidth}{!}{
    \begin{tabular}{*{1}{c}|*{2}{c}|*{2}{c}}
        \toprule[1pt] 
        Method &\textit{R-Ratio} &\textit{R-Start} & mmiou &ldiou \\
        \midrule
         &\checkmark & &42.22 &32.74 \\
        ours-TEA &\checkmark &\checkmark &44.72 &33.36 \\
        \bottomrule 
    \end{tabular}}
    \label{tab:random}
\end{table}

\begin{table}
    \caption{\textbf{Comparison results of distillation losses using different embeddings.} We supervise the student's feature space using cross-entropy loss. Overall, all distillation losses are necessary.}
    \centering
    \resizebox{0.48\textwidth}{!}{
    \begin{tabular}{*{1}{c}|*{3}{c}|*{2}{c}}
        \toprule[1pt] 
        Method &\textit{Tem.} &\textit{Spa.} &\textit{Proto.} & mmIoU &LDIoU \\
        \midrule
         &\checkmark & & &42.87 &31.50 \\
         & &\checkmark & &42.28 &31.10 \\
         &\checkmark &\checkmark & &43.27 &31.62 \\
        ours-TEA &\checkmark &\checkmark &\checkmark &44.72 &33.36 \\
        \bottomrule 
    \end{tabular}}
    \label{tab:teastuema}
\end{table}

\subsection{Experimental Results}
We conducted a comprehensive evaluation on ten sequence lengths from 10\% to 100\%, covering eight temporal image segmentation models, the baseline method we adopted, and our proposed method TEA.

\begin{figure*}
    \centering
    \includegraphics[width=1.0\linewidth]{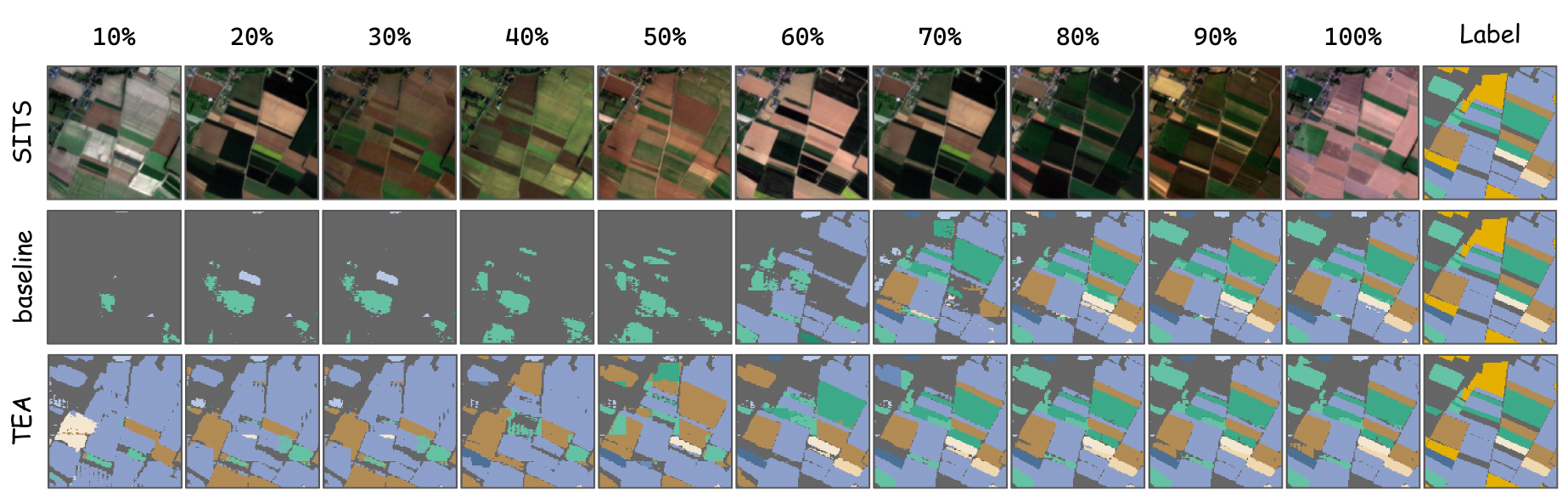}
    \caption{Visualization of segmentation results on PASTIS across different sequence lengths. Baseline denotes the TSViT model.}
    \label{fig:exp}
\end{figure*}

\textbf{Comparative analysis of methods.} 
The essential requirement for a temporally length adaptive model is to maintain strong segmentation performance across image sequences of varying lengths.
We found that the stronger-performing base models across different sequence lengths were mostly derived from U-TAE and our baseline, which integrated a temporal Transformer module. 
We take the best-performing TSViT as our baseline and integrate our proposed method. The results in \cref{tab:main} show that compared with the baseline, in terms of mmIoU, our approach improves performance by 17.61 and 20.56 on the PASTIS and Germany datasets, corresponding to relative gains of 65\% and 53\%, respectively. For our proposed LDIoU metric, our method achieves improvements of 19.28 and 17.72 on the two datasets, reaching roughly a twofold increase. These results indicate that we have effectively enhanced adaptability to sequence length, and that LDIoU highlights the model’s performance on shorter sequences.

\textbf{Results across temporal length ratios.} 
In general, shorter temporal sequences contain less information, making feature extraction and classification more challenging. Consistent with our assumption, our results exhibit performance degradation as the sequence length decreases, which underscores the necessity of our proposed LDIoU. 
Our method outperforms existing approaches across all sequence lengths. 
Notably, on the PASTIS dataset with the full 100\% sequence, our method improves mIoU by 2.7, achieving state-of-the-art (SOTA) performance on the complete sequence. 
This gain is on par with the improvement that TSViT achieved over prior methods. 
We argue that TEA is not only robust to variable sequence lengths but also strengthens the learning of temporal details, thereby delivering gains even on long sequences.

\subsection{Ablation Studies}
In this section, we aim to further demonstrate the superiority of all components or configurations in our approach. 
We conduct ablation experiments to investigate each component separately on the Pastis dataset and comprehensively compare various metrics.

\textbf{All the components matter.}
Our approach comprises a random-cropping data processing strategy and three optimization components: a knowledge distillation framework, a temporal prototype alignment module, and a data reconstruction module. We validate the contribution of each module (see \cref{tab:ablation}). The results show that random-cropping training yields a substantial improvement, raising performance from 14.08 to 28.93 in LDIoU. Building on this, the distillation method alone provides an additional gain of 4.04, while the combination of prototype alignment and data reconstruction delivers a comparable improvement. The distinction lies in that the former enables more effective feature extraction from short sequences, whereas the latter facilitates finer-grained segmentation on long sequences.

When all components are used together, we observe a further gain of 0.4. 
We further conducted ablation studies on the prototype alignment module and the data reconstruction module, and the results indicate that both are effective in improving performance.

\textbf{Random Start also matters.}
For data cropping strategy in the training process, we evaluated two settings: (1) random-length truncation starting from the first frame, and (2) random-length truncation starting from a random position. Results are reported in \cref{tab:random}. We find that when both the ratio and the starting position are randomized, the model learns finer-grained details across different positions in the sequence, leading to more stable segmentation outcomes.

\textbf{Comprehensive feature-space distillation losses work.}
In \cref{tab:teastuema}, we compare three schemes: temporal-only distillation, spatial-only distillation, and temporal-spatio distillation. 
The results indicate that transferring temporal knowledge is more critical than spatial knowledge, while combining both yields the best performance. 
Moreover, aligning temporal prototype knowledge is necessary, providing a 1.74-point gain in LDIoU.

\section{Conclusion}

In this paper, we propose TEA, a temporal adaptive SITS segmentation approach that expands the agricultural application scenarios of image series semantic segmentation models. 
We enhance robustness to varying sequence lengths via proposed modules and knowledge distillation framework.
Extensive experiments and comprehensive evaluations collectively showcase the superiority of our approach.
We believe our work is of significant importance to practical agricultural production. 
If follow-up work can further improve the precision of temporal adaptive models, it will adapt to more downstream tasks and yield significant economic and agriculture value.
{
    \small
    \bibliographystyle{ieeenat_fullname}
    \bibliography{main}

@String(ECCV= {Eur. Conf. Comput. Vis.})

@String(ICLR = {Int. Conf. Learn. Represent.})

@String(AAAI = {AAAI})

@String(ECCV  = {ECCV})

@String(ICLR  = {ICLR})

@article{ConvLSTM,
  title={Convolutional LSTM network: A machine learning approach for precipitation nowcasting},
  author={Shi, Xingjian and Chen, Zhourong and Wang, Hao and Yeung, Dit-Yan and Wong, Wai-Kin and Woo, Wang-chun},
  journal={Advances in neural information processing systems},
  volume={28},
  year={2015}
}

@article{ConvGRU,
  title={Delving deeper into convolutional networks for learning video representations},
  author={Ballas, Nicolas and Yao, Li and Pal, Chris and Courville, Aaron},
  journal={ICLR},
  year={2016}
}

@article{BiConvGRU,
  title={Multi-temporal land cover classification with sequential recurrent encoders},
  author={Ru{\ss}wurm, Marc and K{\"o}rner, Marco},
  journal={ISPRS International Journal of Geo-Information},
  volume={7},
  number={4},
  pages={129},
  year={2018}
}

@inproceedings{BiConvLSTM,
  title={Bidirectional convolutional lstm for the detection of violence in videos},
  author={Hanson, Alex and Pnvr, Koutilya and Krishnagopal, Sanjukta and Davis, Larry},
  booktitle={Proceedings of the European conference on computer vision (ECCV) workshops},
  year={2018}
}

@article{FPNConvLSTM,
  title={Fully convolutional recurrent networks for multidate crop recognition from multitemporal image sequences},
  author={Martinez, Jorge Andres Chamorro and La Rosa, Laura Elena Cu{\'e} and Feitosa, Raul Queiroz and Sanches, Ieda Del’Arco and Happ, Patrick Nigri},
  journal={ISPRS Journal of Photogrammetry and Remote Sensing},
  volume={171},
  pages={188--201},
  year={2021}
}

@misc{lin2017featurepyramidnetworksobject,
      title={Feature Pyramid Networks for Object Detection}, 
      author={Tsung-Yi Lin and Piotr Dollár and Ross Girshick and Kaiming He and Bharath Hariharan and Serge Belongie},
      year={2017},
      eprint={1612.03144},
      archivePrefix={arXiv},
      primaryClass={cs.CV},
      url={https://arxiv.org/abs/1612.03144}, 
}

@inproceedings{UNET3D,
  title={Semantic segmentation of crop type in Africa: A novel dataset and analysis of deep learning methods},
  author={M Rustowicz, Rose and Cheong, Robin and Wang, Lijing and Ermon, Stefano and Burke, Marshall and Lobell, David},
  booktitle={Proceedings of the IEEE/CVF conference on computer vision and pattern recognition workshops},
  pages={75--82},
  year={2019}
}

@article{UNET3Df,
  title={Context-self contrastive pretraining for crop type semantic segmentation},
  author={Tarasiou, Michail and G{\"u}ler, Riza Alp and Zafeiriou, Stefanos},
  journal={IEEE Transactions on Geoscience and Remote Sensing},
  volume={60},
  pages={1--17},
  year={2022}
}

@inproceedings{UTAE,
  title={Panoptic segmentation of satellite image time series with convolutional temporal attention networks},
  author={Garnot, Vivien Sainte Fare and Landrieu, Loic},
  booktitle={Proceedings of the IEEE/CVF International Conference on Computer Vision},
  pages={4872--4881},
  year={2021}
}

@article{germany,
  title={Multi-temporal land cover classification with sequential recurrent encoders},
  author={Ru{\ss}wurm, Marc and K{\"o}rner, Marco},
  journal={ISPRS International Journal of Geo-Information},
  volume={7},
  number={4},
  pages={129},
  year={2018}
}

@inproceedings{TSViT,
  title={Vits for sits: Vision transformers for satellite image time series},
  author={Tarasiou, Michail and Chavez, Erik and Zafeiriou, Stefanos},
  booktitle={Proceedings of the IEEE/CVF Conference on Computer Vision and Pattern Recognition},
  pages={10418--10428},
  year={2023}
}

@inproceedings{zhu2025exact,
  title={Exact: Exploring space-time perceptive clues for weakly supervised satellite image time series semantic segmentation},
  author={Zhu, Hao and Zhu, Yan and Xiao, Jiayu and Xiao, Tianxiang and Ma, Yike and Zhang, Yucheng and Dai, Feng},
  booktitle={Proceedings of the Computer Vision and Pattern Recognition Conference},
  pages={14036--14045},
  year={2025}
}

@article{TempCNN,
  title={Temporal convolutional neural network for the classification of satellite image time series},
  author={Pelletier, Charlotte and Webb, Geoffrey I and Petitjean, Fran{\c{c}}ois},
  journal={Remote Sensing},
  volume={11},
  number={5},
  pages={523},
  year={2019}
}

@article{GLTAE,
  title={Attention to both global and local features: A novel temporal encoder for satellite image time series classification},
  author={Zhang, Weixiong and Zhang, Hao and Zhao, Zhitao and Tang, Ping and Zhang, Zheng},
  journal={Remote Sensing},
  volume={15},
  number={3},
  pages={618},
  year={2023}
}

@article{3DCNN,
  title={3D convolutional neural networks for crop classification with multi-temporal remote sensing images},
  author={Ji, Shunping and Zhang, Chi and Xu, Anjian and Shi, Yun and Duan, Yulin},
  journal={Remote Sensing},
  volume={10},
  number={1},
  pages={75},
  year={2018}
}

@inproceedings{m2019semantic,
  title={Semantic segmentation of crop type in Africa: A novel dataset and analysis of deep learning methods},
  author={M Rustowicz, Rose and Cheong, Robin and Wang, Lijing and Ermon, Stefano and Burke, Marshall and Lobell, David},
  booktitle={Proceedings of the IEEE/cvf conference on computer vision and pattern recognition workshops},
  pages={75--82},
  year={2019}
}

@inproceedings{ronneberger2015u,
  title={U-net: Convolutional networks for biomedical image segmentation},
  author={Ronneberger, Olaf and Fischer, Philipp and Brox, Thomas},
  booktitle={International Conference on Medical image computing and computer-assisted intervention},
  pages={234--241},
  year={2015},
  organization={Springer}
}

@inproceedings{CNNRNN,
  title={Time-space tradeoff in deep learning models for crop classification on satellite multi-spectral image time series},
  author={Garnot, V Sainte Fare and Landrieu, Loic and Giordano, Sebastien and Chehata, Nesrine},
  booktitle={IEEE International Geoscience and Remote Sensing Symposium},
  pages={6247--6250},
  year={2019}
}

@article{End-to-end,
  title={End-to-end learned early classification of time series for in-season crop type mapping},
  author={Ru{\ss}wurm, Marc and Courty, Nicolas and Emonet, R{\'e}mi and Lef{\`e}vre, S{\'e}bastien and Tuia, Devis and Tavenard, Romain},
  journal={ISPRS Journal of Photogrammetry and Remote Sensing},
  volume={196},
  pages={445--456},
  year={2023},
  publisher={Elsevier}
}

@article{schafer2020teaser,
  title={TEASER: early and accurate time series classification},
  author={Sch{\"a}fer, Patrick and Leser, Ulf},
  journal={Data mining and knowledge discovery},
  volume={34},
  number={5},
  pages={1336--1362},
  year={2020},
  publisher={Springer}
}

@inproceedings{ghalwash2014utilizing,
  title={Utilizing temporal patterns for estimating uncertainty in interpretable early decision making},
  author={Ghalwash, Mohamed F and Radosavljevic, Vladan and Obradovic, Zoran},
  booktitle={Proceedings of the 20th ACM SIGKDD international conference on Knowledge discovery and data mining},
  pages={402--411},
  year={2014}
}

@inproceedings{grabocka2014learning,
  title={Learning time-series shapelets},
  author={Grabocka, Josif and Schilling, Nicolas and Wistuba, Martin and Schmidt-Thieme, Lars},
  booktitle={Proceedings of the 20th ACM SIGKDD international conference on Knowledge discovery and data mining},
  pages={392--401},
  year={2014}
}

@inproceedings{garnot2020satellite,
  title={Satellite image time series classification with pixel-set encoders and temporal self-attention},
  author={Garnot, Vivien Sainte Fare and Landrieu, Loic and Giordano, Sebastien and Chehata, Nesrine},
  booktitle={Proceedings of the IEEE/CVF Conference on Computer Vision and Pattern Recognition},
  pages={12325--12334},
  year={2020}
}

@inproceedings{buciluǎ2006model,
  title={Model compression},
  author={Buciluǎ, Cristian and Caruana, Rich and Niculescu-Mizil, Alexandru},
  booktitle={Proceedings of the 12th ACM SIGKDD international conference on Knowledge discovery and data mining},
  pages={535--541},
  year={2006}
}

@article{hinton2015distilling,
  title={Distilling the knowledge in a neural network},
  author={Hinton, Geoffrey and Vinyals, Oriol and Dean, Jeff},
  journal={arXiv preprint arXiv:1503.02531},
  year={2015}
}

@article{tarvainen2017mean,
  title={Mean teachers are better role models: Weight-averaged consistency targets improve semi-supervised deep learning results},
  author={Tarvainen, Antti and Valpola, Harri},
  journal={Advances in neural information processing systems},
  volume={30},
  year={2017}
}

@article{wang2018dataset,
  title={Dataset distillation},
  author={Wang, Tongzhou and Zhu, Jun-Yan and Torralba, Antonio and Efros, Alexei A},
  journal={arXiv preprint arXiv:1811.10959},
  year={2018}
}

@inproceedings{ahn2019variational,
  title={Variational information distillation for knowledge transfer},
  author={Ahn, Sungsoo and Hu, Shell Xu and Damianou, Andreas and Lawrence, Neil D and Dai, Zhenwen},
  booktitle={Proceedings of the IEEE/CVF conference on computer vision and pattern recognition},
  pages={9163--9171},
  year={2019}
}

@inproceedings{aguilar2020knowledge,
  title={Knowledge distillation from internal representations},
  author={Aguilar, Gustavo and Ling, Yuan and Zhang, Yu and Yao, Benjamin and Fan, Xing and Guo, Chenlei},
  booktitle={Proceedings of the AAAI conference on artificial intelligence},
  volume={34},
  number={05},
  pages={7350--7357},
  year={2020}
}

@article{allen2019learning,
  title={Learning and generalization in overparameterized neural networks, going beyond two layers},
  author={Allen-Zhu, Zeyuan and Li, Yuanzhi and Liang, Yingyu},
  journal={Advances in neural information processing systems},
  volume={32},
  year={2019}
}

@inproceedings{bergmann2020uninformed,
  title={Uninformed students: Student-teacher anomaly detection with discriminative latent embeddings},
  author={Bergmann, Paul and Fauser, Michael and Sattlegger, David and Steger, Carsten},
  booktitle={Proceedings of the IEEE/CVF conference on computer vision and pattern recognition},
  pages={4183--4192},
  year={2020}
}

@inproceedings{Omnisat,
  title={Omnisat: Self-supervised modality fusion for earth observation},
  author={Astruc, Guillaume and Gonthier, Nicolas and Mallet, Clement and Landrieu, Loic},
  booktitle={European Conference on Computer Vision},
  pages={409--427},
  year={2024}
}

@article{S2,
  title={Sentinel-2: ESA's optical high-resolution mission for GMES operational services},
  author={Drusch, Matthias and Del Bello, Umberto and Carlier, S{\'e}bastien and Colin, Olivier and Fernandez, Veronica and Gascon, Ferran and Hoersch, Bianca and Isola, Claudia and Laberinti, Paolo and Martimort, Philippe and others},
  journal={Remote sensing of Environment},
  volume={120},
  pages={25--36},
  year={2012}
}

@article{S1,
  title={GMES Sentinel-1 mission},
  author={Torres, Ramon and Snoeij, Paul and Geudtner, Dirk and Bibby, David and Davidson, Malcolm and Attema, Evert and Potin, Pierre and Rommen, Bj{\"O}rn and Floury, Nicolas and Brown, Mike and others},
  journal={Remote sensing of environment},
  volume={120},
  pages={9--24},
  year={2012}
}

@article{woodcock2008free,
  title={Free access to Landsat imagery.},
  author={Woodcock, Curtis E and Allen, Richard and Anderson, Martha and Belward, Alan and Bindschadler, Robert and Cohen, Warren and Gao, Feng and Goward, Samuel N and Helder, Dennis and Helmer, Eileen and others},
  journal={Science},
  year={2008}
}

@inproceedings{benson2024multi,
  title={Multi-modal learning for geospatial vegetation forecasting},
  author={Benson, Vitus and Robin, Claire and Requena-Mesa, Christian and Alonso, Lazaro and Carvalhais, Nuno and Cort{\'e}s, Jos{\'e} and Gao, Zhihan and Linscheid, Nora and Weynants, M{\'e}lanie and Reichstein, Markus},
  booktitle={Proceedings of the IEEE/CVF Conference on Computer Vision and Pattern Recognition},
  pages={27788--27799},
  year={2024}
}

@article{Satmae,
  title={Satmae: Pre-training transformers for temporal and multi-spectral satellite imagery},
  author={Cong, Yezhen and Khanna, Samar and Meng, Chenlin and Liu, Patrick and Rozi, Erik and He, Yutong and Burke, Marshall and Lobell, David and Ermon, Stefano},
  journal={Advances in Neural Information Processing Systems},
  volume={35},
  pages={197--211},
  year={2022}
}

@inproceedings{Skysense,
  title={Skysense: A multi-modal remote sensing foundation model towards universal interpretation for earth observation imagery},
  author={Guo, Xin and Lao, Jiangwei and Dang, Bo and Zhang, Yingying and Yu, Lei and Ru, Lixiang and Zhong, Liheng and Huang, Ziyuan and Wu, Kang and Hu, Dingxiang and others},
  booktitle={Proceedings of the IEEE/CVF Conference on Computer Vision and Pattern Recognition},
  pages={27672--27683},
  year={2024}
}

@inproceedings{S2mae,
  title={S2mae: A spatial-spectral pretraining foundation model for spectral remote sensing data},
  author={Li, Xuyang and Hong, Danfeng and Chanussot, Jocelyn},
  booktitle={Proceedings of the IEEE/CVF Conference on Computer Vision and Pattern Recognition},
  pages={24088--24097},
  year={2024}
}

@article{improvement,
  title={Improvement in crop mapping from satellite image time series by effectively supervising deep neural networks},
  author={Mohammadi, Sina and Belgiu, Mariana and Stein, Alfred},
  journal={ISPRS Journal of Photogrammetry and Remote Sensing},
  volume={198},
  pages={272--283},
  year={2023}
}

@article{zhong2019deep,
  title={Deep learning based multi-temporal crop classification},
  author={Zhong, Liheng and Hu, Lina and Zhou, Hang},
  journal={Remote sensing of environment},
  volume={221},
  pages={430--443},
  year={2019}
}

@inproceedings{yang2018robust,
  title={Robust classification with convolutional prototype learning},
  author={Yang, Hong-Ming and Zhang, Xu-Yao and Yin, Fei and Liu, Cheng-Lin},
  booktitle={Proceedings of the IEEE conference on computer vision and pattern recognition},
  pages={3474--3482},
  year={2018}
}

@inproceedings{li2021adaptive,
  title={Adaptive prototype learning and allocation for few-shot segmentation},
  author={Li, Gen and Jampani, Varun and Sevilla-Lara, Laura and Sun, Deqing and Kim, Jonghyun and Kim, Joongkyu},
  booktitle={Proceedings of the IEEE/CVF conference on computer vision and pattern recognition},
  pages={8334--8343},
  year={2021}
}

@article{sun2024learning,
  title={Learning local and global temporal contexts for video semantic segmentation},
  author={Sun, Guolei and Liu, Yun and Ding, Henghui and Wu, Min and Van Gool, Luc},
  journal={IEEE Transactions on Pattern Analysis and Machine Intelligence},
  volume={46},
  number={10},
  pages={6919--6934},
  year={2024},
  publisher={IEEE}
}

@inproceedings{yarats2021reinforcement,
  title={Reinforcement learning with prototypical representations},
  author={Yarats, Denis and Fergus, Rob and Lazaric, Alessandro and Pinto, Lerrel},
  booktitle={International Conference on Machine Learning},
  pages={11920--11931},
  year={2021},
  organization={PMLR}
}

@article{loshchilov2017decoupled,
  title={Decoupled weight decay regularization},
  author={Loshchilov, Ilya and Hutter, Frank},
  journal={arXiv preprint arXiv:1711.05101},
  year={2017}
}

@article{loshchilov2016sgdr,
  title={Sgdr: Stochastic gradient descent with warm restarts},
  author={Loshchilov, Ilya and Hutter, Frank},
  journal={arXiv preprint arXiv:1608.03983},
  year={2016}
}
}
\clearpage
\setcounter{page}{1}
\maketitlesupplementary

In this supplementary material, we mainly present the following contents: 1) additional experimental details and test results analyzation; 2) research on adapting our method to the U-TAE baseline; 3) visualizations of the model’s results from the feature layers to the output layer, which comprehensively demonstrate the advantages of our method.

\section{Additional Implementation Details}
\label{sec:rationale}
\subsection{Datasets Preprocessing}
\textbf{Temporal Encoding.} Due to the strict temporal order requirement of our task, we fine-tuned the original dataset to align with our task objectives.
The Pastis dataset has a temporal resolution of 5 days, with a time span of approximately 400 days, ranging from August of the previous year to October of the following year. We modified the setting in TSViT where encoding is based on the absolute day of the year, and instead adopted the first sampling date of the dataset as the start time, using the difference between the actual date and the start time as the temporal encoding. 

This encoding method aligns the temporal order of remote sensing images with the real crop growth laws and ensures the early-stage characteristics of the data after cropping.
\textbf{Zero padding.} Since the original dataset removed some low-quality data with severe issues such as cloud occlusion, resulting in inconsistent data lengths ranging from 38 to 61 samples per sequence, we performed zero-padding for the missing dates to pad each sample to 80 frames, which ensures that the cropped samples have the same time span.
For the Germany dataset, we adopt the same preprocessing method, with the padded length being 46. Since most samples have zero-padding in the last few images of the sequence, which has almost no impact on the overall dataset performance evaluation, we consider only retaining the first 36 images to form our dataset. 


\subsection{Additional Test Results}
Earth observation data is typically continuous. Thanks to satellites’ fixed revisit cycles and stable data acquisition capabilities, people can generally obtain satellite image time series starting from sowing at any stage of crop growth. Based on this assumption, we adopted the approach of cropping from the sequence start to present the test results in the main text, which is also more practically applicable. However, considering some extreme and special cases—such as the loss of remote sensing image sequences caused by satellite technical failures or long-term extreme weather—we supplemented test results of sequences with different starting positions and lengths to comprehensively demonstrate the \textbf{temporal adaptability} of our model.

Here, we tested the results of sliding along the temporal dimension with four lengths (10\%, 20\%, 40\%, and 80\%) at a step size of 10\%. (For example, the \textit{\textbf{start frame}} of 80\% and the \textbf{\textit{sequence length }} of 10\% means extracting the segment from 80\% to 90\% of the original sequence, while the \textit{\textbf{start frame}} of 80\% and the \textbf{\textit{sequence length }} of 20\% presents 80\% to 100\%.) Our results (shown in \cref{tab:seq_1}) comprehensively demonstrate its ability to extract information from short-term time series, from which we can draw the following conclusions from the results:

\begin{table*}
    \caption{\textbf{Performance Comparison Under Multiple Sequence Cropping Modes.} We conducted tests on the Pastis and Germany datasets, with the same methods and parameter settings as those in the main text. The first row presents the baseline (TSViT) results, while the second row, which is formatted in bold, shows the results of TEA.}
    \centering
    \resizebox{\textwidth}{!}{%
    \begin{tabular}{*{1}{c}|*{1}{c}|*{10}{c}}
        \toprule[1pt] 
         \multicolumn{2}{c|}{\multirow{2}{*}{\textit{PASTIS}}} &\multicolumn{10}{c}{start frame} \\
         \multicolumn{2}{c|}{}& 0\% & 10\% & 20\% & 30\% & 40\% & 50\% & 60\% & 70\% & 80\% & 90\%  \\
          \midrule
          \multirow{10}{*}{\rotatebox{90}{sequence length }} 
          &\multirow{2}{*}{10\%}  &3.81 &3.07 &2.34 &2.83 &4.77 &5.20 &8.43 &20.54 &8.38 &3.65   \\
          & &\textbf{21.5} &\textbf{21.35} &\textbf{16.28} &\textbf{27.28} &\textbf{33.55} &\textbf{40.94} &\textbf{50.31} &\textbf{58.78} &\textbf{42.28} &\textbf{32.34} \\
          \cmidrule{2-12}
          &\multirow{2}{*}{20\%} &5.60 &3.48 &3.01 &6.18 &12.76 &18.48 &40.63 &30.11 &15.51 &-  \\
          & &\textbf{26.22} &\textbf{26.06} &\textbf{29.04} &\textbf{34.93} &\textbf{44.21} &\textbf{54.48} &\textbf{63.44} &\textbf{60.60} &\textbf{44.79} &-  \\
          \cmidrule{2-12}
          &\multirow{2}{*}{40\%} &6.48 &7.89 &15.39 &28.11 &52.55 &57.38 &51.44 &- &- &-  \\
          & &\textbf{32.70} &\textbf{36.84} &\textbf{44.81} &\textbf{55.71} &\textbf{65.51} &\textbf{65.76} &\textbf{65.37} &- &- &-  \\
          \cmidrule{2-12}
          &\multirow{2}{*}{60\%} &20.34 &31.07 &53.65 &60.04 &61.64 &- &- &- &- &-  \\
          & &\textbf{45.82} &\textbf{56.17} &\textbf{65.30} &\textbf{66.29} &\textbf{66.58} &- &- &- &- &-  \\
          \cmidrule{2-12}
          &\multirow{2}{*}{80\%} &56.46 & 61.41 &62.28 &- &- &- &- &- &- &-  \\
          & &\textbf{65.36} &\textbf{66.34} &\textbf{66.57} &- &- &- &- &- &- &-  \\
         \bottomrule[1pt] 
    \end{tabular}
    }
    \resizebox{\textwidth}{!}{%
    \begin{tabular}{*{1}{c}|*{1}{c}|*{10}{c}}
        \toprule[1pt] 
         \multicolumn{2}{c|}{\multirow{2}{*}{\textit{Germany}}} &\multicolumn{10}{c}{start frame} \\
         \multicolumn{2}{c|}{}& 0\% & 10\% & 20\% & 30\% & 40\% & 50\% & 60\% & 70\% & 80\% & 90\%  \\
          \midrule
          \multirow{10}{*}{\rotatebox{90}{sequence length }} 
          &\multirow{2}{*}{10\%}  &2.42 &3.71 &2.56 &3.39 &17.01 &4.03 &26.03 &10.40 &12.53 &7.31   \\
          & &\textbf{2.49} &\textbf{20.61} &\textbf{21.60} &\textbf{27.32} &\textbf{54.70} &\textbf{29.95} &\textbf{72.07} &\textbf{46.80} &\textbf{46.40} &\textbf{30.53}  \\
          \cmidrule{2-12}
          &\multirow{2}{*}{20\%} &4.62 &5.37 &3.91 &18.73 &22.76 &26.50 &37.24 &14.97 &14.41 &-  \\
          & &\textbf{30.25} &\textbf{34.62} &\textbf{40.89} &\textbf{58.76} &\textbf{67.25} &\textbf{75.66} &\textbf{78.77} &\textbf{57.82} &\textbf{49.41} &-  \\
          \cmidrule{2-12}
          &\multirow{2}{*}{40\%} &5.55 &24.44 &27.01 &58.42 &69.87 &44.08 &46.77 &- &- &-  \\
          & &\textbf{46.92} &\textbf{62.91} &\textbf{70.84} &\textbf{82.66} &\textbf{84.80} &\textbf{81.51} &\textbf{81.22} &- &- &-  \\
          \cmidrule{2-12}
          &\multirow{2}{*}{60\%} &30.58 &65.78 &75.28 &79.42 &79.94 &- &- &- &- &-  \\
          & &\textbf{72.20} &\textbf{83.65} &\textbf{85.52} &\textbf{85.84} &\textbf{85.63} &- &- &- &- &-  \\
          \cmidrule{2-12}
          &\multirow{2}{*}{80\%} &79.22 & 83.61 &84.10 &- &- &- &- &- &- &-  \\
          & &\textbf{85.69} &\textbf{86.18} &\textbf{86.06} &- &- &- &- &- &- &-  \\
         \bottomrule[1pt] 
    \end{tabular}
    }
    \label{tab:seq_1}
\end{table*}

\begin{itemize}
    \item For the Pastis dataset, the results incorporating 70\%-80\% of the sequence are relatively better, while for the Germany dataset, the optimal performance is achieved when including 60\%-70\% of the sequence. We attribute this to the higher crop maturity during this period, which enables remote sensing images to contain richer crop information and more distinct feature differences, thereby facilitating better differentiation.
\end{itemize}

\begin{itemize}
    \item Overall, when cropping sequences by sliding along the time axis with the same length, the performance of crop parcel segmentation shows a trend of first increasing and then decreasing. This aligns with the practical law that the distinguishability of crops is positively correlated with their maturity during natural growth. Thus, segmenting crops only using early growth stages is not only necessary but also highly challenging, which further highlights the significance of our research on the temporal adaptive model and the superiority of our method.
\end{itemize}

\begin{itemize}
    \item Compared with the baseline, TEA achieves a comprehensive and significant improvement in the basic metrics. With just \textbf{10\%} of the sequence, TEA reaches \textbf{88\%} of the performance delivered by the complete sequence—highlighting its exceptional ability to leverage limited temporal data.
\end{itemize}

\begin{table*}
    \caption{Various temporal length performance of different models (including method improved upon U-TAE baseline, denoted as \textbf{TEA(U-TAE)} ) on PASTIS and Germany. The best results are highlighted in \textbf{bold}, while the second-best results are marked in \underline{underline}. (* denotes the introduction of an FPN head.)}
    \centering
    \resizebox{\textwidth}{!}{%
    \begin{tabular}{*{1}{c}|*{1}{c}|*{10}{c}|*{1}{c}|*{1}{c}}
        \toprule[1pt] 
        \multirow{2}{*}{ }  & \multirow{2}{*}{Method}   & \multicolumn{10}{c|}{ratios}  &mmIoU  & LDIoU   \\
                    & &10\% &20\% &30\% &40\% &50\% &60\% &70\% &80\% &90\% &100\% &(\%) &(\%) \\
        
        \midrule 
        \multirow{10}{*}{\rotatebox{90}{\textit{\textbf{PASTIS }}}}  
        &UNet3D  & 3.23 & 4.17 & 4.77 & 5.86 & 6.34 & 11.18 & 19.83 & 39.14 & 47.32 & 50.84 & 19.27 & 10.10 \\
        &UNet3Df  & 3.63 & 3.72 & 3.74 & 4.95 & 6.91 & 9.66 & 20.58 & 33.50 & 44.59 & 49.58 & 18.09 & 9.56 \\
        &ConvGRU  & 2.40 & 3.94 & 4.28 & 4.82 & 5.36 & 7.26 & 7.24 & 15.59 & 28.74 & 50.57 & 13.02 & 7.00 \\
        &BiConvGRU   &2.29 &3.83 &4.13 &5.38 &7.45 &9.42 &15.59 &25.15 &24.39 &46.42 &14.40 &7.75  \\
        &ConvLSTM  & 3.28 & 3.34 & 3.47 & 3.58 & 3.89 & 4.23 & 5.17 & 7.85 & 22.11 & 51.67 & 10.86 & 6.09 \\
        &ConvLSTM*  & 0.36 & 0.38 & 0.53 & 1.06 & 1.45 & 0.74 & 0.55 & 2.00 & 29.00 & 55.90 & 9.20 & 3.60 \\
        &BiConvLSTM  &5.51 &6.19 &7.53 &9.07 &9.31 &9.85 &13.78 &21.84 &29.72 &46.50 &15.93 &10.08  \\
        \cmidrule{2-14}
        &U-TAE  & 3.15 & 5.93 & 6.34 & 8.67 & 11.19 & 21.52 & 33.17 & 52.72 & 60.55 & 61.40 & 26.46 & 13.80 \\
        &TEA(U-TAE) &\underline{10.96} &\underline{14.50} &\underline{17.12} &\underline{25.13} &\underline{30.57} &\underline{39.73} &\underline{51.17} &\underline{61.59} &\underline{62.87} & 63.09 &\underline{37.67} &\underline{24.32} \\
        \cmidrule{2-14}
        &TSViT & 3.81 & 5.60 & 6.10 & 6.48 &11.2 &20.34 &34.42 &56.46 &62.65 &\underline{64.08} &27.11 &14.08 \\
        &TEA(TSViT) &\textbf{21.5} &\textbf{26.22} &\textbf{28.43} &\textbf{32.70} &\textbf{37.57} &\textbf{45.82} &\textbf{56.45} &\textbf{65.36} &\textbf{66.37} &\textbf{66.77} &\textbf{44.72} &\textbf{33.36} \\
        
        \midrule 
        \multirow{10}{*}{\rotatebox{90}{\textit{\textbf{Germany }}}} 
        &UNet3D & 2.44 & 2.48 & 2.43 & 3.01 & 11.94 & 16.93 & 27.05 & 32.52 & 65.79 & 73.66 & 23.83 & 11.29 \\
        &UNet3Df & \textbf{2.65} & 3.12 & 3.57 & 3.42 & 15.19 & 15.56 & 30.04 & 39.73 & 65.44 & 74.32 & 25.30 & 12.24 \\
        &ConvGRU & 1.32 & 2.43 & 2.32 & 2.29 & 3.04 & 3.09 & 4.80 & 5.03 & 11.20 & 59.00 & 9.45 & 4.60 \\
        &BiConvGRU   &2.22 &3.03 &3.01 &3.09 &4.03 &3.61 &3.15 &4.67 &6.7 &54.99 &8.85 &4.85  \\
        &ConvLSTM & 0.62 & 1.26 & 1.45 & 1.34 & 3.08 & 2.16 & 6.39 & 10.57 & 26.68 & 63.15 & 11.67 & 4.97 \\
        &ConvLSTM* & 0.58 & 0.53 & 0.58 & 1.93 & 14.64 & 17.41 & 27.65 & 41.85 & 64.60 & 73.19 & 24.30 & 10.59 \\
        &BiConvLSTM  &2.03 &2.16 &1.83 &1.14 &1.49 &1.34 &1.28 &2.49 &6.89 &58.1 &7.88 &3.96  \\
        \cmidrule{2-14}
        &U-TAE & 2.44 & 3.37 & 3.26 & 7.69 & 20.80 & 26.30 & 56.55 & 75.87 & 79.39 & 82.00 & 35.77 & 17.16 \\
        &TEA(U-TAE) &2.44 &\underline{17.44} &\underline{17.57} &\underline{28.99} &\underline{48.37} &\underline{53.36} &\underline{71.19} &75.26 &75.70 &75.30 &\underline{46.56} &\underline{26.75} \\
        \cmidrule{2-14}
        &TSViT &2.42 &4.62 &4.90 &5.55 &24.74 &30.58 &67.57 &\underline{79.22} &\underline{84.30} &\underline{86.28} &39.02 & 18.90 \\
        &TEA(ours) &\underline{2.49} &\textbf{30.25} &\textbf{34.64} &\textbf{46.92} &\textbf{66.87} &\textbf{72.20} &\textbf{84.18} &\textbf{85.69} &\textbf{86.24} &\textbf{86.36} &\textbf{59.58} &\textbf{36.62} \\
        \bottomrule[1pt] 
    \end{tabular}
    }
    \label{tab:utae_tea}
\end{table*}
\section{Generalization Analysis of Method}
In the main text, considering that the TSViT model achieves the best performance in existing relevant studies and serves as a universally recognized performance benchmark for this task, our method is developed with TSViT as the core baseline to conduct targeted improvements, ensuring the comparability and advancement of technical optimizations. 
Notably, through preliminary exploratory experiments, we observe that although the UTAE model does not reach the same performance as TSViT in the field, its comprehensive performance falls within the same level as TSViT and significantly outperforms other counterpart baseline models, demonstrating substantial potential as a backbone for further enhancements. Based on this finding, to further validate the generalizability of the proposed technical framework, we deeply integrate our core methodology with the UTAE model architecture, designing and implementing a novel model variant.

Specifically, this variant still adheres to the core training paradigm of knowledge distillation: utilizing randomly cropped temporal sequences as input data, constructing supervision signals by leveraging full-temporal knowledge transferred from the teacher model, and guiding the student model to efficiently learn key information and intrinsic correlations of temporal features. Meanwhile, to strengthen the model’s capability in modeling temporal dependencies and the robustness of feature representations, we organically embed the temporal prototype alignment module and the data reconstruction module into the original UTAE network architecture. Through the synergistic effect of these two modules, the global consistency and local detailed information of temporal data are further explored.

Experimental results presented in \cref{tab:utae_tea} demonstrate that all performance metrics of the fused model are significantly superior to those of various baseline methods, including TSViT and the original UTAE. This verifies that the proposed training framework and core modules are not confined to specific baseline models but possess excellent cross-model adaptability and generalization performance. 

When comparing the results of our TEA method under the two baseline settings, we find that the TSViT-based approach achieves a more significant improvement than the UTAE-based one. We attribute this discrepancy mainly to the following reasons: 
\begin{itemize}
    \item The temporal feature extraction module LTAE of UTAE adopts a more lightweight attention mechanism compared to the Temporal Transformer of TSViT, where only the keys are learned through linear layers. Therefore, by incorporating richer learnable parameters and stacking more layers, TSViT effectively expands the richness of temporal feature extraction and elevates the upper bound of the model's knowledge learning for variable-length time series.
\end{itemize} 
\begin{itemize}
    \item Additionally, TSViT is designed with k class tokens (where k equals the number of classes) to learn temporal features, preserving more effective information compared to UTAE which globally pools the sequence into a single channel. 
\end{itemize}
Under the full-temporal training and testing of the baselines, due to the rich input information and distinct feature differences, the performance gap between these two architectures is relatively small; however, in our method based on TSViT exhibits better flexibility and richer semantic information, leading to a more substantial improvement over its baseline than UTAE.

In future work, the optimization of our method can focus on two core directions: first, further enhancing the basic feature extraction capability of the base model and exploring network architecture designs more suitable for temporal tasks; second, deepening the dynamic adjustment mechanism of the temporal adaptive framework to improve the model’s adaptability to different temporal scales and data distribution scenarios.

\begin{figure*}
    \centering
    \includegraphics[width=0.9\linewidth]{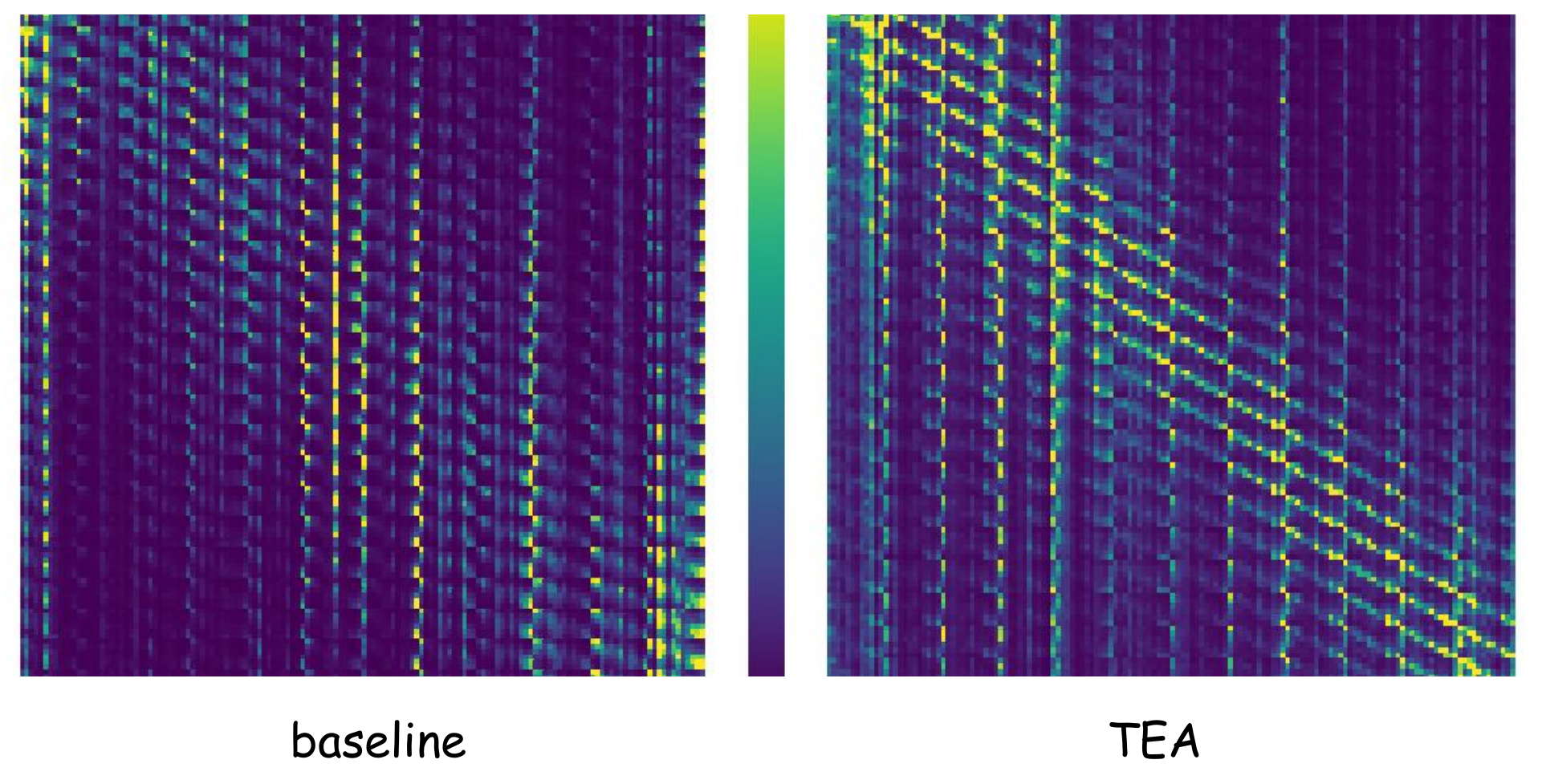}
    \caption{\textbf{Attention Weight Visualization} of spatial. Both the horizontal and vertical axes represent the number of tokens. The color intensity represents the activation magnitude, with yellow at the top indicating the maximum value and blue at the bottom indicating the minimum value.}
    \label{fig:cam}
\end{figure*}

\section{Visualization}
\subsection{Attention Visualization}
We generate Attention Weight Visualization plots based on the output of the last feature layer of the Transformer network. 
As illustrated in \cref{fig:cam}, our method results in denser attention among tokens, showing a diagonal trend—meaning they pay greater attention to the features of themselves and neighboring tokens. 
In contrast, the original TSViT method exhibits sparser attention, focusing primarily on individual tokens; some tokens even activate all positions. Through analysis, we attribute this phenomenon to the global knowledge accumulation enabled by random cropping and knowledge distillation. Specifically, the original method only classifies by selecting a few tokens with strong discriminative features, whereas our method, benefiting from random cropping, can learn knowledge from all tokens in a balanced manner. Consequently, the method reduces reliance on key features and can comprehensively learn knowledge from all token features, achieving more robust segmentation performance.

\subsection{segmentation}
Here, we present additional segmentation results to highlight the superiority of our method from a practical application perspective, as well as to demonstrate its significance and value for agricultural production activities.

\begin{figure*}
    \centering
    \includegraphics[width=1.0\linewidth]{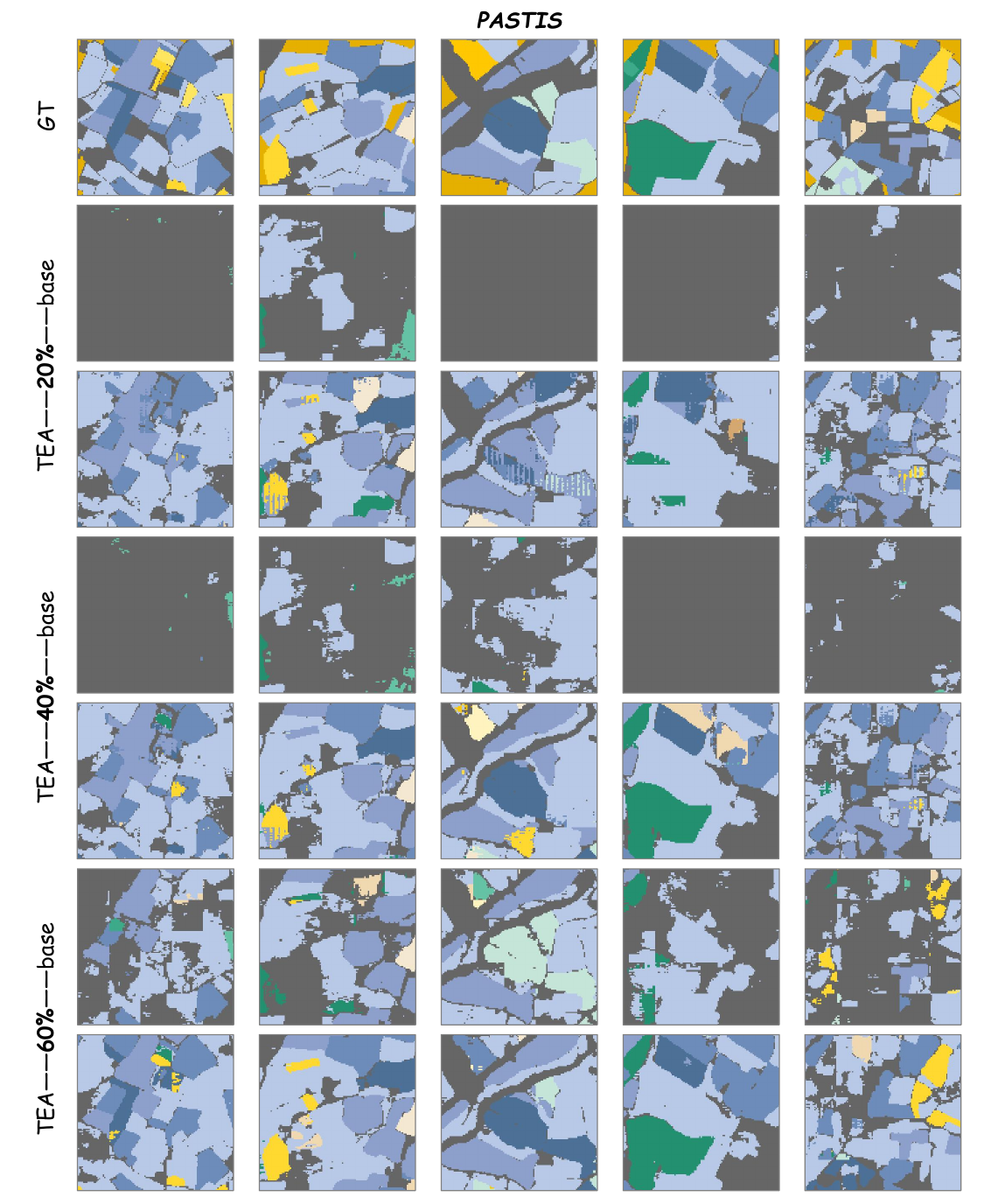}
    \label{fig:visual-pastis}
\end{figure*}
\begin{figure*}
    \centering
    \includegraphics[width=1.0\linewidth]{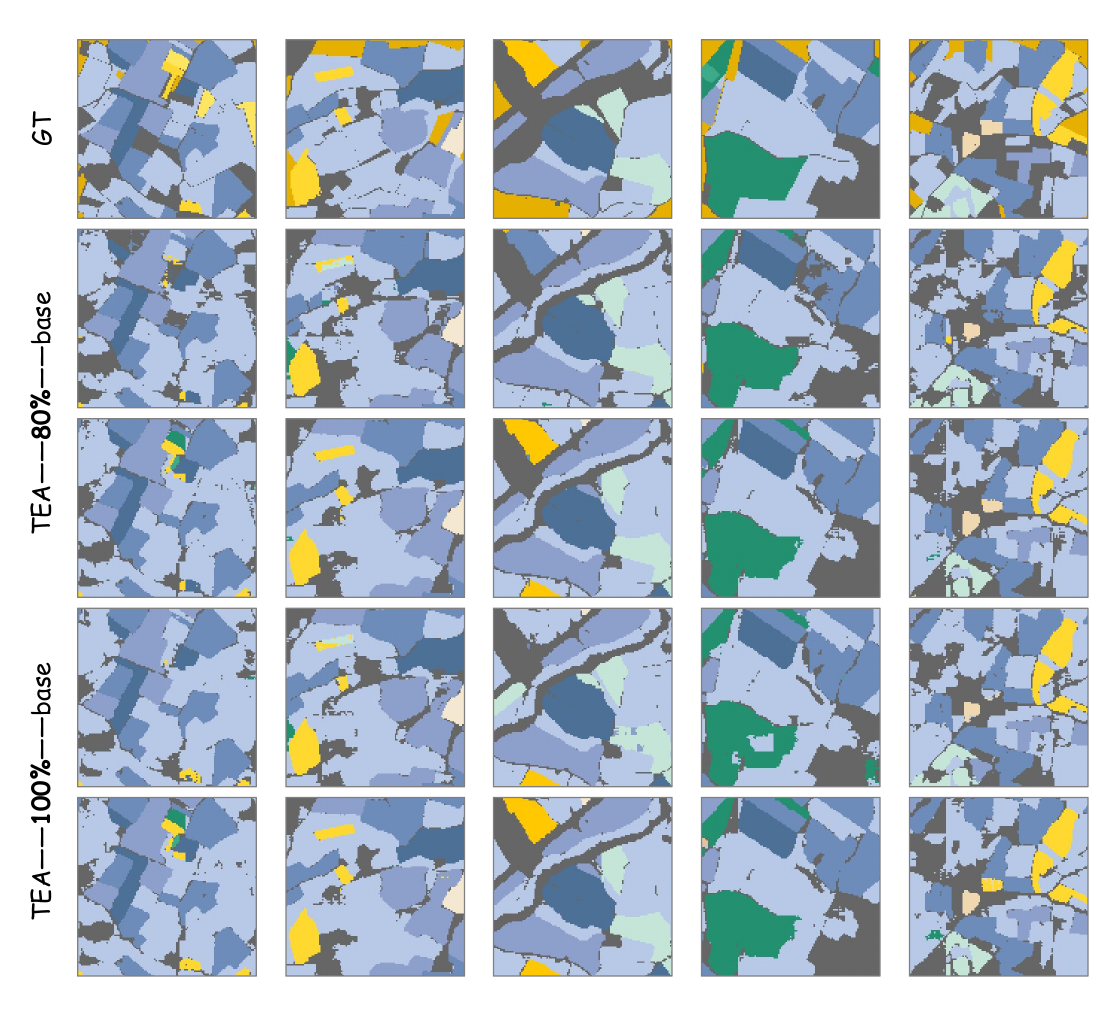}
    \caption{\textbf{Qualitative comparison between the baseline method and our TEA method on the PASTIS dataset.} We mainly present the short-term time series testing results cropped at ratios of 20\%, 40\%, and 60\% from the start of the sequence. The model input during testing is the same as during training, with a size of 24×24. We recombine the images of the same region into a 120×120 size image according to the cropping indices generated during data preprocessing.}
    \label{fig:visual-pastis}
\end{figure*}

\begin{figure*}
    \centering
    \includegraphics[width=1.0\linewidth]{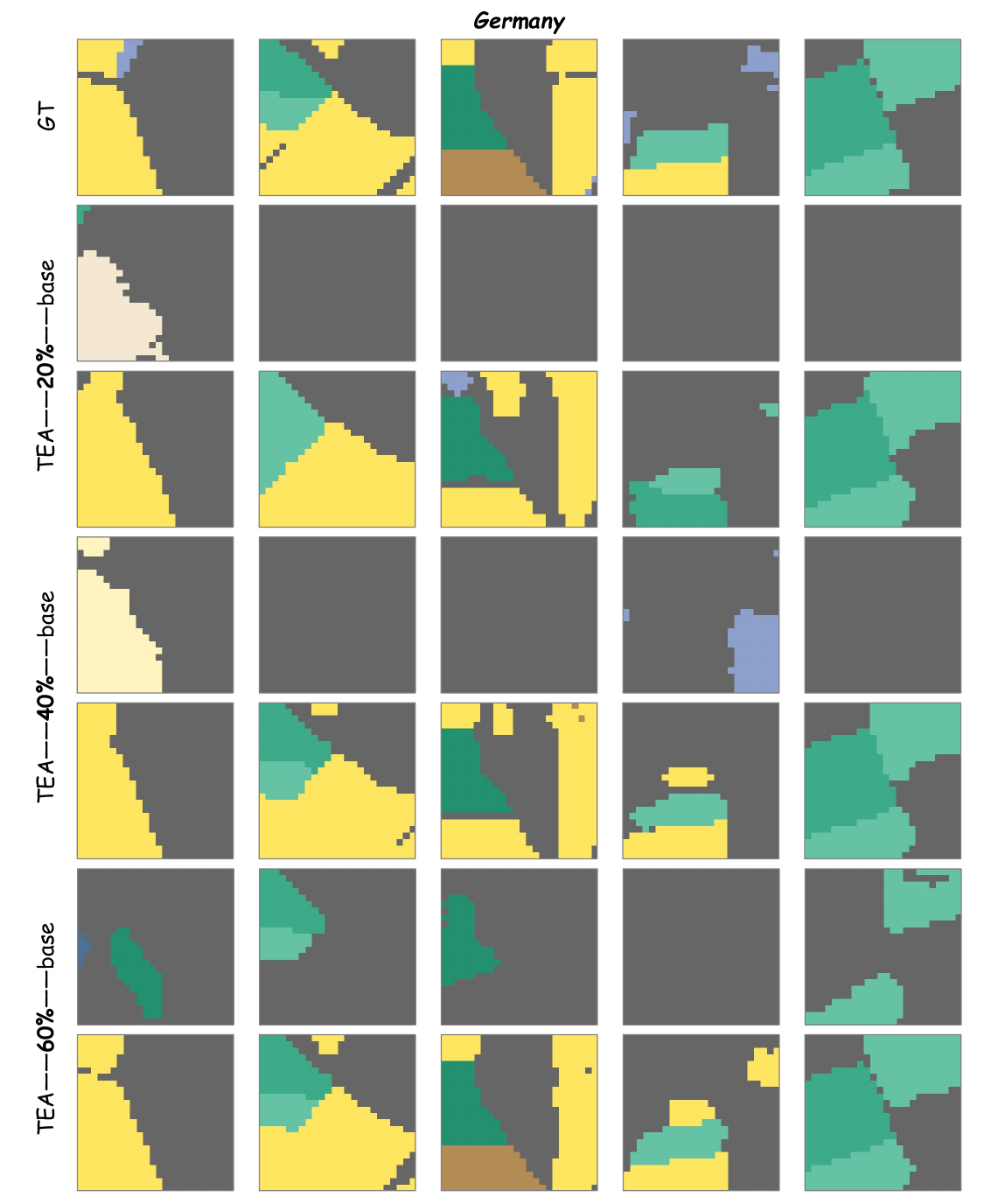}
    \label{fig:visual-mtlcc}
\end{figure*}
\begin{figure*}
    \centering
    \includegraphics[width=1.0\linewidth]{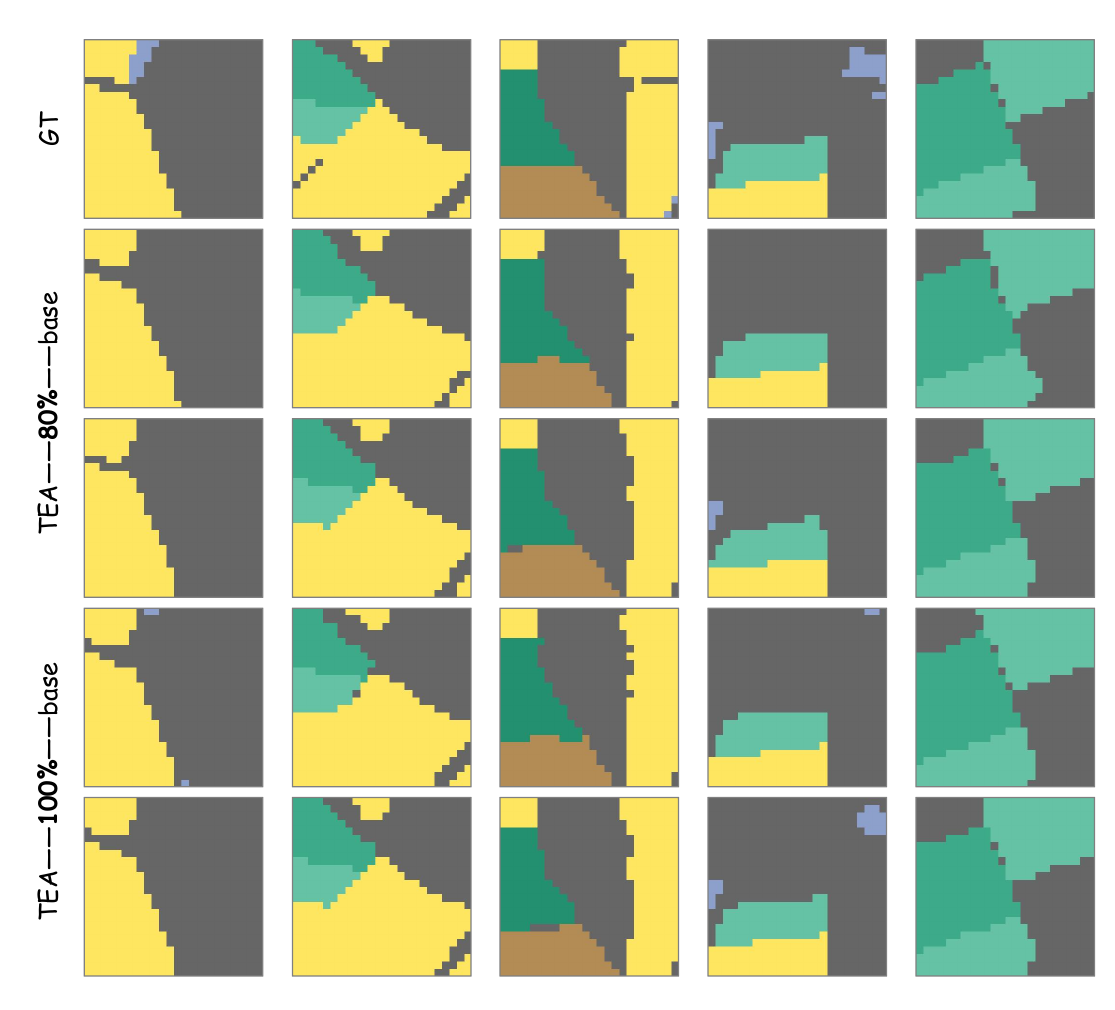}
    \caption{\textbf{Qualitative comparison between the baseline method and our TEA method on the Germany dataset.} We mainly present the short-term time series testing results cropped at ratios of 20\%, 40\%, and 60\% from the start of the sequence. The Germany dataset consists of images with a size of 24×24.}
    \label{fig:visual-mtlcc}
\end{figure*}



\end{document}